\documentclass[11pt]{report}

\usepackage[numbers]{natbib}
\usepackage{amsmath}
\usepackage{algorithm}
\usepackage{algpseudocode}
\usepackage{amssymb}
\usepackage{booktabs}
\usepackage[left=3cm,right=3cm,top=3cm,bottom=3cm]{geometry}
\usepackage{adjustbox}
\usepackage{url}
\usepackage{pgfplots}
\pgfplotsset{compat=1.18}
\usepackage{subcaption}
\usetikzlibrary{arrows}
\usepackage{multirow}
\usepackage[pdftex,pagebackref,hypertexnames=false,colorlinks]{hyperref} 
\hypersetup{
    colorlinks=true,       
    linkcolor=black,        
    citecolor=blue,        
    filecolor=magenta,     
    urlcolor=blue          
}

\title{Robust Time Series Causal Discovery for Agent-Based Model Validation}

\author{Gene Yu, Ce Guo, and Wayne Luk\\
\{gene.yu23, c.guo, w.luk\}@imperial.ac.uk\\
Department of Computing\\
Imperial College London, UK
}
\date{}

\begin{document}
\maketitle

\begin{abstract}
Agent-Based Model (ABM) validation is crucial as it helps ensuring the reliability of simulations, and causal discovery has become a powerful tool in this context. However, current causal discovery methods often face accuracy and robustness challenges when applied to complex and noisy time series data, which is typical in ABM scenarios. This study addresses these issues by proposing a Robust Cross-Validation (RCV) approach to enhance causal structure learning for ABM validation. We develop RCV-VarLiNGAM and RCV-PCMCI, novel extensions of two prominent causal discovery algorithms. These aim to reduce the impact of noise better and give more reliable causal relation results, even with high-dimensional, time-dependent data. The proposed approach is then integrated into an enhanced ABM validation framework, which is designed to handle diverse data and model structures.
\\ \\
The approach is evaluated using synthetic datasets and a complex simulated fMRI dataset. The results demonstrate greater reliability in causal structure identification. The study examines how various characteristics of datasets affect the performance of established causal discovery methods. These characteristics include linearity, noise distribution, stationarity, and causal structure density. This analysis is then extended to the RCV method to see how it compares in these different situations. This examination helps confirm whether the results are consistent with existing literature and also reveals the strengths and weaknesses of the novel approaches.
\\ \\
By tackling key methodological challenges, the study aims to enhance ABM validation with a more resilient valuation framework presented. These improvements increase the reliability of model-driven decision making processes in complex systems analysis.
\end{abstract}

\tableofcontents 

\chapter{Introduction}

\section{Agent-Based Models: An Overview}
In the modern era, Agent-Based Models (ABMs) fall under the class of modelling and simulation techniques which are increasingly being used in domains such as theoretical economics, finance, social sciences, and epidemiology. 
It involves a bottom-up approach by putting together individual agents and their interactions so that various phenomena may be synthesised and then examined. The detailed methodologies of such models allow them to understand complexity and draw futuristic conclusions.
\\ \\
Recent literature stresses the significant advantages of agent-based models (ABMs) over traditional economic models. Fagiolo (2019) pinpoints the major strengths of ABMs: their capacity to provide comprehensive narratives of interactions among agents with network structures, incomplete information learning processes, and competition in imperfect markets—and the flexibility they provide in validating both model inputs and outputs \cite{Fagiolo2019}. This characteristic has gained much attention and prompted much research activity recently.

\section{The Importance of ABM Validation}
To successfully implement ABMs in reality, the ``validation" of this model could be the decisive factor for its ability to truly reflect real-world reality. The validation process consists of comparing the model output with the actual data obtained in the real world to make sure it is reliable and effective. This process is utilised to establish the credibility of the model and the reliability of the predictions made.
\\ \\
Validation is particularly challenging in fields with complex interactions and non-linear dynamics, such as finance. As Windrum et al. (2007) pointed out, significant effort is still needed to realise consistent and satisfactory techniques of ABM method implementation to real-world financial data~\cite{Windrum2007}. 
\\ \\
A key component in ABM validation is to find the cause-and-effect mechanisms from data. Besides highlighting the importance of correlational testing, causal matching between the ABM outputs and real-world data has recently been emphasized in validation. These approaches aim to understand and explain the origins and propagation of observed phenomena in financial systems \cite{Guerini2017}. The details of such causal discovery methods and their application in ABM validation will be further discussed in the following chapters.

\section{Challenges to Address}
Indeed, in spite of continuous progress in the ABM validation techniques, there are still the most significant challenges in such complex systems applications:
\\
\begin{enumerate}
    \item \textbf{Insufficient Robustness of Time Series Causal Discovery Methods}:
    The causal discovery approaches that are commonly used today, such as VAR-LiNGAM and PCMCI, can be quite susceptible to noise and variations in the data. This could create inconsistencies when the method is applied to different subsets of the same dataset or datasets with slightly different characteristics. For ABM validation purposes, the robustness of these techniques should be enhanced. False causality in this regard could be as harmful as wrong conclusions drawn up about the ABM system's validity and, consequently, about the underlying system mechanisms.

    \item \textbf{Lack of Comprehensive Understanding of Dataset Characteristics' Impact}:
    While previous studies have examined specific dataset characteristics, there is a lack of comprehensive understanding of how various dataset properties collectively affect the performance of causal discovery methods in ABM validation. The absence of a systematic comparison across a wide range of dataset types (e.g., linear vs. non-linear, Gaussian vs. non-Gaussian noise, stationary vs. non-stationary, sparse vs. dense causal structures) hinders our ability to select appropriate validation techniques and interpret results accurately. This deficiency stands as a huge constraint in the area of creating functional ABM validation processes as a solution to the wide spectrum of complicated problems.
    
    \item \textbf{Limitations in Existing ABM Validation Frameworks}: Even though much progress has been made in ABM Validation, for instance, the framework proposed by Guerini et al. (2017) \cite{Guerini2017}, current approaches still face several key limitations: 

    \begin{enumerate}
        \item[(a)] Insufficient Dataset Property Analysis: Existing frameworks often lack comprehensive tests for some important dataset properties. For instance, some of these may overlook important characteristics such as linearity and stationarity, which are essential for understanding the nature of the data and selecting appropriate modelling techniques.

        \item[(b)] Limited Options for Causal Discovery Methods: Most current frameworks rely on a single or limited set of causal discovery methods. This limitation may prevent us from obtaining optimal performance when dealing with different dataset characteristics or different priorities (such as accuracy vs. efficiency). The lack of method diversity limits the framework's adaptability to various scenarios and data types.

        \item[(c)] Narrow Range of Performance Metrics: The existing validation framework typically focuses on basic similarity tests or a limited set of performance metrics. This may cause to failure to capture the full range of model performance, especially in complex financial systems where causal relationships can be intricate.
    \end{enumerate}
\end{enumerate}

Addressing these challenges is crucial for advancing ABM validation in complex systems like financial markets. Overcoming these limitations will lead to more reliable models, enhancing decision-making processes and informing policy decisions. These challenges necessitate innovative approaches in causal discovery and model validation techniques applicable to a wide range of complex systems.

\section{Novel Approaches and Contributions}
This research addresses the challenges mentioned above by utilizing several new approaches and providing the following key contributions:
\\
\begin{enumerate}
    \item \textbf{Robust Cross-Validated (RCV) Causal Discovery Method (Chapter 3)}:
    We introduce a novel approach to enhance the robustness of existing causal discovery methods. By applying cross-validation techniques to causal discovery algorithms such as VAR-LiNGAM and PCMCI, we aim to mitigate the sensitivity of these methods to noise and data variations to improve the consistency and reliability of causal structure identification in complex time series data. 

    \item \textbf{Comprehensive Experimental Evaluation and Analysis (Chapter 4)}:
    We present a thorough empirical analysis of our proposed methods and existing approaches. This evaluation covers a wide range of characteristics, including linear vs. non-linear relationships, Gaussian vs. non-Gaussian noise, stationary vs. non-stationary behaviour, and sparse vs. dense causal structures. We also examine the scalability of methods with varying numbers of variables and time series lengths. We ran experimental evaluations on both synthetic datasets with controlled properties and a complex simulated fMRI dataset, so as to provide insights into method performance under various conditions.
    
    \item \textbf{Context-Aware ABM Validation Framework (Chapter 5)}:
    Extending the work done by Guerini, we develop an enhanced ABM validation framework that addresses the weaknesses of existing one. In this framework, the user could choose a suitable method of causal inference based on the property of data sets or other validation needs, such as efficiency or accuracy. Another improvement we feature is to include a more comprehensive set of performance metrics for assessing the causal relations to get a more precise evaluation of model performance. This framework is designed to pre-process datasets and ensure their uniformity, analyse dataset attributes, run user-dependent or driven causal structure detection, and enhance validation evaluations.
\end{enumerate}

These contributions enhance ABM validation in complex systems by improving causal discovery methods and offering a flexible validation framework. This research aims to increase the accuracy and reliability of ABMs in capturing real-world dynamics across various domains.

\section{Report Structure}
The remainder of this report is structured as follows: Chapter 2 provides the foundational background and reviews related work, covering Agent-Based Models in finance, existing ABM validation techniques, causal discovery methods and the current limitations. Chapter 3 introduces our novel Robust Cross-Validated (RCV) Causal Discovery Approach, detailing its theoretical foundations and implementation. Chapter 4 presents our experimental evaluation and analysis, including synthetic dataset generation, comparative analysis of causal discovery methods, and an application to a complex simulated fMRI dataset that mimics real-world neuroimaging data. Chapter 5 describes our Context-Aware ABM Validation Framework, explaining how it integrates improved causal discovery methods and enhances overall validation reliability. Finally, Chapter 6 concludes the report, summarizing key findings, discussing implications for ABM validation in complex systems and suggesting future research directions.

\chapter{Background and Related Work}

\section{Agent-Based Models in Finance}
Agent-Based Models (ABMs) are increasingly being employed for simulating systems' complexity, specifically in finance and economics. LeBaron (2000) underscores the involvement of such models in thinking dynamics of economic models' interactions as he states it is: ``beginning to show promise as a research methodology that will greatly impact how we think about interactions in economic models" \cite{LeBaron2000}. The basic principle of ABM is based on the behaviour that each agent adapts to the conditions that affect the market. They are especially conducive to examining complex phenomena, those emergent market behaviors, such as price movements, can be generated by simple rules of interaction at the agent level.
\\ \\
The ABMs turned out to be particularly helpful in explaining the economic crisis and thus making way for the implementation of good policy measures. They could illustrate how individual interactions can lead to emerging economic patterns \cite{Farmer2009} or be used to study wealth dynamics and economic inequality \cite{Impullitti2002}. Raberto, et al. (2012) took a step further, exploring other arenas besides borrowing and the debt of borrowers to find out how they valued economic production. Their research suggests that such structural models can demonstrate the potential to depict complex economic relationships by keeping the complex relationship between the physical and financial variables of the economy. \cite{Raberto2012}.
\\ \\
Recently, interest in highly accurate market simulation as a supportive tool for AI research into financial applications has increased. ABIDES (Agent-Based Interactive Discrete Event Simulation) is a platform tailored as a research space for AI agents used in financial markets, as shown in Byrd (2020) \cite{Byrd2020}. These platforms can provide researchers with the toolbox for simulating intricate market dynamics and applying AI algorithms within the context of realistic financial cases.

\section{ABM Validation Approaches}
Validation of Agent-Based Models (ABMs) is critical to ensure they accurately represent real-world systems and provide reliable insights \cite{Law2009}. The validation process typically involves several key dimensions. Ziegler categorized validity into three different types \cite{Zeigler2018}:

\begin{itemize}
    \item \textbf{Replicative Validity}: Ensuring that the model fits external data.
    \item \textbf{Predictive Validity}: Ensuring that the model matches data that can be acquired from the modelled system.
    \item \textbf{Structural Validity}: Ensuring that the model accurately reflects the processes it is designed to simulate.
\end{itemize}

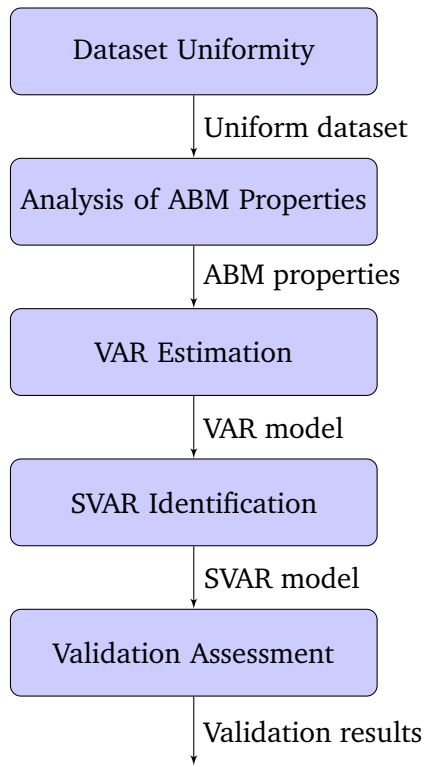
\begin{figure}[h!]
    \centering
    \begin{tikzpicture}[node distance=2cm]

        \tikzstyle{block} = [rectangle, draw, fill=blue!20, 
            text width=12em, text centered, rounded corners, minimum height=3em]
        \tikzstyle{line} = [draw, -latex']

        \node [block] (step1) {Dataset Uniformity};
        \node [block, below of=step1] (step2) {Analysis of ABM Properties};
        \node [block, below of=step2] (step3) {VAR Estimation};
        \node [block, below of=step3] (step4) {SVAR Identification};
        \node [block, below of=step4] (step5) {Validation Assessment};

        \path [line] (step1) -- node [midway, right] {Uniform dataset} (step2);
        \path [line] (step2) -- node [midway, right] {ABM properties} (step3);
        \path [line] (step3) -- node [midway, right] {VAR model} (step4);
        \path [line] (step4) -- node [midway, right] {SVAR model} (step5);
        \path [line] (step5) -- ++(0, -1.5) node [midway, right, align=left] {Validation results};

    \end{tikzpicture}
    \caption{The five steps of the validation method, adapted from Guerini and Moneta (2017) \cite{Guerini2017}.}
    \label{fig:five-step}
\end{figure}

These concepts match the definition of validation stated in the literature, which emphasises the importance of demonstrating that a model correctly reflects the real world for its intended use \cite{Cooley2017, Macal2005}. 
\\ \\
In the context of financial ABMs, validation techniques have evolved to address the complex dynamics of financial markets. Traditional methods often focus on comparing model outputs with historical data, but recent progress has led to more advanced approaches.
\\ \\
One of the important developments in ABM validation is the causal matching approach proposed by Guerini and Moneta \cite{Guerini2017}. This approach consists of five steps as shown in Figure~\ref{fig:five-step}: ensuring dataset consistency, analyzing ABM characteristics, estimating Vector Autoregression (VAR) models, identifying Structural VAR (SVAR), and assessing the validation \cite{Guerini2017}. This method aims to reproduce the causal relations observed in real financial systems, providing a more rigorous validation framework that goes beyond simple correlational matching.
\\

Other validation techniques include:
\begin{itemize}
    \item \textbf{Indirect Calibration:} This method compares ABM simulation results with real-world data, focusing on reproducing key statistical properties or stylized facts about financial markets \cite{Winker2007}.
    
    \item \textbf{Method of Simulated Moments:} This approach estimates model parameters by matching moments of simulated data to moments of observed data \cite{Gilli2003}.
    
    \item \textbf{Bayesian Estimation:} This technique uses Bayesian inference to estimate model parameters that can incorporate prior knowledge and uncertainty quantification \cite{Grazzini2017}.
    
    \item \textbf{Pattern-Oriented Modelling:} This approach uses many patterns observed in real-world systems to guide the development and validation of models to capture various aspects of financial systems accurately \cite{Grimm2005}.
\end{itemize}

While these methods have shown promise, challenges remain in validating complex financial ABMs. These include handling the high dimensionality of financial data, capturing non-linear relationships, and accounting for the often non-stationary characteristics of financial time series.
\\ \\
Recent work has highlighted the importance of integrating parameter calibration, estimation, and space exploration into the empirical validation of ABMs in economics and finance \cite{Fagiolo2019}. This integrated approach allows for a more thorough validation process that considers multiple aspects of model performance.
\\ \\
As the field evolves, more studies are trying to use advanced statistical and machine-learning techniques for ABM validation. These developments aim to improve the robustness and reliability of ABMs, thereby enhancing their usefulness for understanding complex systems and informing policy decisions.

\section{Causal Graphs in Time Series}

\begin{figure}[htbp]
\centering
\begin{subfigure}[b]{0.48\textwidth}
\centering
\adjustbox{height=7cm,keepaspectratio}{%
\includegraphics[width=\textwidth]{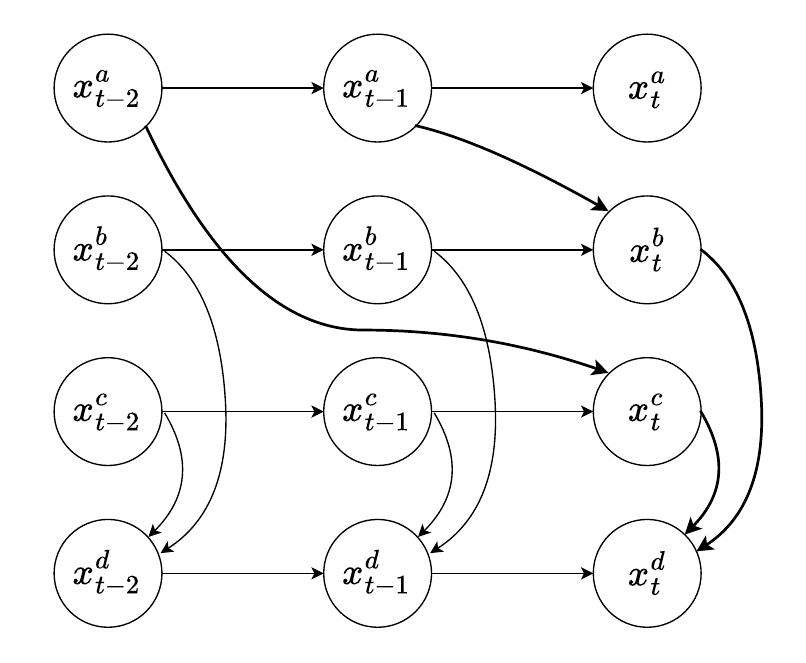}
}
\caption{Window Causal Graph}
\label{fig:window_graph}
\end{subfigure}
\hfill
\begin{subfigure}[b]{0.48\textwidth}
\centering
\adjustbox{height=7cm,keepaspectratio}{%
\includegraphics[width=\textwidth]{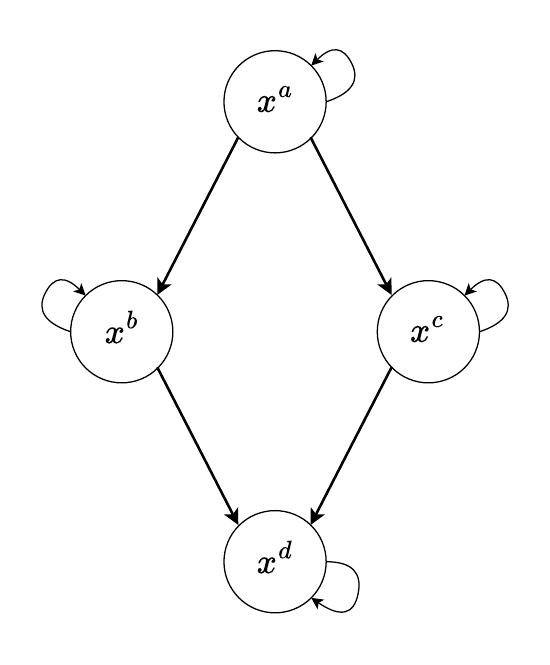}
}
\caption{Summary Causal Graph}
\label{fig:summary_graph}
\end{subfigure}
\caption{Graphical representations of causal relationships in time series data. (a) The window causal graph depicts interactions within a defined time frame, capturing temporal dynamics. (b) The summary causal graph provides a condensed overview of causal relationships, abstracting from specific time points. While (a) offers more temporal detail, (b) presents a simplified, atemporal view of the overall causal structure. (Reproduced from Assaad et al. \cite{Assaad2022})}
\label{fig:causal_graphs}
\end{figure}

Before examining specific causal discovery methods, it is important to understand how causal relationships in time series data are represented graphically. While full-time causal graphs provide a comprehensive view of causal structures over an entire time series, modern causal discovery methods typically focus on inferring two more practical representations: window causal graphs and summary causal graphs. Figure \ref{fig:causal_graphs} illustrates these two graphical representations.
\\ \\
Window causal graphs focus on causal relations within a specific time window. They typically include a fixed number of lags and provide a local view of the causal structure.
\\ \\
Summary causal graphs condense the causal relations into a single, time-independent representation. They provide an overview of the causal structure. They retain only the relevant causal relations between variables, without taking into account the time lags. 
\\ \\
These graphical representations serve as a foundation for understanding and analyzing causal relations in time series data. This is particularly important in ABM validation, as it enables researchers to visualize and quantify the similarities and differences in causal structures between ABM simulated data and real-world data.
\\ \\
Understanding these representations is vital as we continue to delve into specific causal discovery methods. These methods aim to infer causal structures from observed time series data, making them a key component in validating ABMs for complex systems.

\section{Causal Discovery Methods}
Causal discovery methods are of great interest in the context of ABM validation, particularly because of their ability to infer causal relations from observational time series data. These methods aim to identify the underlying causal structure of a system, which is important for understanding the mechanisms behind observed phenomena in financial markets.

\subsection{VAR-LiNGAM (Linear Non-Gaussian Acyclic Model)}
Vector Autoregressive (VAR) models serve as a foundation for many causal discovery methods in time series analysis. VAR models capture linear interdependencies among multiple time series, expressing each variable as a linear function of past values of itself and past values of other variables \cite{Lutkepohl2005}. The general form of a VAR model of order $p$, denoted as VAR($p$), is:

\begin{equation}
    \mathbf{X}_t = \mathbf{c} + \mathbf{A}_1\mathbf{X}_{t-1} + \mathbf{A}_2\mathbf{X}_{t-2} + \cdots + \mathbf{A}_p\mathbf{X}_{t-p} + \boldsymbol{\varepsilon}_t
    \label{eq:var_model}
\end{equation}

where $\mathbf{X}_t$ is a $k \times 1$ vector of variables at time $t$, $\mathbf{c}$ is a $k \times 1$ vector of constants, $\mathbf{A}_i$ are $k \times k$ coefficient matrices, and $\boldsymbol{\varepsilon}_t$ is a $k \times 1$ vector of error terms.
\\ \\
While VAR models do not determine causality, they provide a basis for more advanced causal discovery techniques. VAR-LiNGAM incorporates the basic VAR model with the Linear Non-Gaussian Acyclic Model (LiNGAM) to identify causal structures in non-Gaussian time series data \cite{Hyvarinen2010}, as shown in Algorithm \ref{alg:var-lingam}. This method is particularly useful in finance data where non-Gaussian distributions are common.

\begin{algorithm}
\caption{VAR-LiNGAM Algorithm}
\label{alg:var-lingam}
\begin{algorithmic}[1]
\Require Time series data $X = \{x_t\}_{t=1}^T$, maximum lag $\tau$
\Ensure Causal structure $G$, adjacency matrices $\{A_k\}_{k=0}^\tau$

\State Estimate VAR model: $x_t = \sum_{k=1}^\tau A_k x_{t-k} + e_t$
\State Obtain residuals $e_t$ from VAR model
\State Apply LiNGAM to residuals $e_t$:
    \State \quad Estimate mixing matrix $B$
    \State \quad Permute $B$ to make it as close to lower triangular as possible
    \State \quad Estimate $A_0 = I - B^{-1}$
\State Compute other adjacency matrices: $A_k = (I - A_0)M_k$ for $k = 1, \ldots, \tau$
\State Construct causal graph $G$ based on $\{A_k\}_{k=0}^\tau$
\State \Return $G$, $\{A_k\}_{k=0}^\tau$
\end{algorithmic}
\end{algorithm}

The key insight of VAR-LiNGAM is that it can determine the instantaneous causal order of variables when the error terms are non-Gaussian, which is not possible with standard VAR models. The VAR-LiNGAM model can be represented as:

\begin{equation}
    \mathbf{X}_t = \mathbf{B}_0\mathbf{X}_t + \sum_{k=1}^\tau \mathbf{B}_k\mathbf{X}_{t-k} + \boldsymbol{\varepsilon}_t
    \label{eq:var_lingam}
\end{equation}

where $\mathbf{B}_0$ is a strictly lower triangular matrix representing contemporaneous causal effects, and $\mathbf{B}_k$ ($k>0$) are unrestricted matrices representing lagged effects.
\\ \\
Recent developments have made VAR-LiNGAM more accessible through open-source Python packages, facilitating its application in various settings including time series cases, multiple-group cases, and hidden common cause scenarios \cite{Ikeuchi2020}.

\subsection{PCMCI (Peter and Clark algorithm with Momentary Conditional Independence)}
PCMCI is a hybrid method that combines the PC algorithm with the concept of momentary conditional independence \cite{Runge2019}, as detailed steps shown in Algorithm \ref{alg:pcmci}. This method is designed to estimate causal graphs from large-scale time series datasets and to address the limitations of traditional methods in handling high-dimensional and nonlinear relationships.

\begin{algorithm}
\caption{PCMCI Algorithm}
\label{alg:pcmci}
\begin{algorithmic}[1]
\Require Time series dataset $\mathbf{X} = (X^1, ..., X^N)$, maximum time lag $\tau_{max}$, significance level $\alpha_{PC}$, $\alpha_{MCI}$
\Ensure Causal graph $\mathcal{G}$

\State // PC step (condition-selection)
\For{each variable $X^j$}
    \State $\hat{\mathcal{P}}(X^j_t) \gets \emptyset$ \Comment{Initialize parents set}
    \State $p \gets 0$ \Comment{Initialize conditioning set size}
    \Repeat
        \For{$X^i_{t-\tau} \in \mathbf{X}^{-}_t \setminus \hat{\mathcal{P}}(X^j_t)$}
            \If{$X^i_{t-\tau} \not\!\perp\!\!\!\perp X^j_t \mid \mathbf{S}$ for any $\mathbf{S} \subseteq \hat{\mathcal{P}}(X^j_t)$ with $|\mathbf{S}|=p$ at $\alpha_{PC}$}
                \State $\hat{\mathcal{P}}(X^j_t) \gets \hat{\mathcal{P}}(X^j_t) \cup \{X^i_{t-\tau}\}$
            \EndIf
        \EndFor
        \State $p \gets p + 1$
    \Until{$p > |\hat{\mathcal{P}}(X^j_t)|$ or $p > p_{max}$}
\EndFor

\State // MCI step
\For{all $X^i_{t-\tau}, X^j_t$}
    \State $\mathcal{P}^{i\rightarrow j}_\tau \gets$ MCI test $X^i_{t-\tau} \not\!\perp\!\!\!\perp X^j_t \mid \hat{\mathcal{P}}(X^j_t) \setminus \{X^i_{t-\tau}\}, \hat{\mathcal{P}}(X^i_{t-\tau})$ at $\alpha_{MCI}$
    \If{$\mathcal{P}^{i\rightarrow j}_\tau < \alpha_{MCI}$}
        \State Add link $X^i_{t-\tau} \rightarrow X^j_t$ to $\mathcal{G}$
    \EndIf
\EndFor

\State \Return $\mathcal{G}$
\end{algorithmic}
\end{algorithm}

The PCMCI algorithm consists of two main steps:
\begin{enumerate}
    \item PC step: This step uses conditional independence tests to identify each variable's parents, reducing the problem's dimensionality.

    \item MCI step: This step uses momentary conditional independence tests to estimate the causal links while controlling for autocorrelation and indirect links.
\end{enumerate}

PCMCI is particularly valuable for its flexibility in handling both linear and nonlinear relationships, making it well-suited for the complex dynamics often observed in financial systems.

\subsection{Other Notable Methods}
While VAR-LiNGAM and PCMCI are the focus of this study, other notable causal discovery methods include:

\begin{itemize}
    \item Granger Causality: Despite its simplicity compared to more recent methods, Granger causality remains a fundamental tool for initial causal analysis in financial time series \cite{Granger1969}. It's based on the principle that if a variable X Granger-causes Y, then past values of X should contain information that helps predict Y beyond the information contained in past values of Y alone.
    
    \item tsFCI (Time Series Fast Causal Inference): An extension of the FCI algorithm adapted for time series data, tsFCI can handle hidden confounders and non-stationary time series \cite{Entner2010}. It's particularly useful in financial contexts where unobserved factors often influence observable variables.
    
    \item DYNOTEARS (Dynamic NOTEARS): This method extends the NOTEARS (Non-combinatorial Optimization via Trace Exponential and Augmented lagRangian for Structure learning) algorithm to time series data \cite{Pamfil2020}. DYNOTEARS can discover both contemporaneous and lagged causal relationships, making it suitable for analyzing complex financial dynamics.
    
    \item CCM (Convergent Cross Mapping): Although not a traditional method for causal discovery, CCM is used to detect causality in nonlinear dynamical systems, which can be particularly relevant for financial markets \cite{Sugihara2012}. It's based on the idea that if X causes Y, then historical values of Y can be used to estimate the states of X.
    
    \item Transfer Entropy: An information-theoretic measure introduced by Schreiber (2000) that quantifies the directed flow of information between two time series. Its ability to distinguish between actual information exchange and shared information due to common history makes it particularly useful for analyzing the interdependence of multivariate time series data \cite{Schreiber2000}.
\end{itemize}

\section{Current Limitations}
While causal discovery methods have significantly advanced ABM validation techniques, several limitations and challenges persist, corresponding to the key challenges identified in Chapter 1:

\subsection{Limitations in Robustness of Causal Discovery Methods}

\begin{itemize}
    \item Sensitivity to Noise and Data Variations: Existing causal discovery methods can be sensitive to noise and variations in the data. This sensitivity can lead to inconsistent results if applied to different subsets of the same dataset or datasets with slightly different characteristics. This instability reduces the reliability of ABM validation and makes it difficult to determine whether the discovered causal relationships reflect underlying economic or financial mechanisms \cite{Runge2019}.

    \item Handling Complex Data: Financial time series often exhibit non-linear relationships, non-Gaussian distributions, and non-stationary behaviour. Although some methods like VAR-LiNGAM and PCMCI can handle some aspects of these complexities, fully capturing the complexity of financial data remains a challenge \cite{Hyvarinen2010, Runge2019}.
\end{itemize}

These limitations will be addressed in Chapter 3, where we introduce our Robust Cross-Validated (RCV) causal discovery approach.

\subsection{Gaps in Understanding Dataset Characteristics' Impact}

\begin{itemize}
    \item Impact of Dataset Characteristics: The performance of causal discovery methods can vary significantly depending on dataset characteristics such as linearity, Gaussianity, stationarity, and the density of causal structures. However, our understanding of how these characteristics influence the performance of causal discovery methods in the context of ABM validation is limited \cite{Assaad2022}.
\end{itemize}

Chapter 4 will present a comprehensive analysis of how various dataset characteristics affect the performance of causal discovery methods, including our proposed RCV approach.

\subsection{Shortcomings of Existing ABM Validation Frameworks}
Existing ABM validation frameworks face several challenges in properly assessing complex models:

\begin{itemize}
    \item Challenges in Capturing Complex Emergent Dynamics: Many current validation approaches have difficulty measuring the fit of simulated results to real-world data. This is particularly true for models with complex and emergent behaviours, such as those found in financial systems. This limitation can lead to incomplete or inaccurate assessment of model validity \cite{Lamperti2018}.

    \item Limited Adaptability in Causal Discovery: While existing frameworks like that of Guerini and Moneta (2017) \cite{Guerini2017} have made significant progress in integrating causal discovery methods, much work remains to be done in terms of method diversity and practical implementation. The ability to flexibly apply and compare multiple causal discovery techniques within a single framework is still not fully realised in many existing approaches. This may limit the framework's effectiveness across the wide range of data characteristics and model structures found in complex financial systems.
\end{itemize}

Our enhanced ABM validation framework, presented in Chapter 5, addresses these limitations by incorporating advanced causal discovery techniques and providing a more flexible approach to model assessment across various scales and complexities.

\subsection{Other Challenges}

\begin{itemize}
    \item Computational Demands: Many causal discovery methods require substantial computing power, particularly when dealing with complex, non-linear connections in large datasets. This can restrict their use in intricate ABMs or extensive financial data analysis \cite{Shimizu2006}. The computational burden often increases with model complexity and data volume, this poses challenges for real-time or large-scale analyses \cite{Hagedorn2022, Hagedorn2023, Guo2022, Guo2023}.
\end{itemize}
While computational efficiency is not the primary focus of our research, we consider this aspect and discuss potential optimisations and trade-offs in Chapters 4 and 5.
\\ \\
Our research aims to tackle these limitations. By doing so, we hope to make causal discovery methods more reliable and applicable in ABM validation. This is especially important when dealing with complex financial systems, where accurate modelling can have significant real-world implications.

\chapter{Proposed Approach}

In this chapter, we present the first major contribution of this study, the Robust Cross-Validated (RCV) Causal Discovery Method. We propose a novel approach that focuses on reinforcing the robustness of existing causal discovery methods, specifically VAR-LiNGAM and PCMCI, by utilizing cross-validation to yield more reliable results. It is this approach that tries to make causal structure deducing more consistent and unbiased, addressing the challenge of sensitivity to noise and data variations in current causal discovery approaches.

\section{Current Limitations}
Modern cause-and-effect discovery techniques face several big problems when used in complicated systems, especially in finance, weather, and brain science. This section outlines the key limitations that our proposed Robust Cross-Validated (RCV) method aims to address, as well as other important challenges in the field.

\subsection{Key Limitations Addressed by RCV}

\begin{itemize}
    \item \textbf{Sensitivity to noise:} Many causal discovery algorithms, quite often those employing conditional independence tests, are very sensitive to noise in the observational data. This sensitivity causes the causal relations to be volatile and incapable of yielding reliable results.

    \item \textbf{Difficulty in handling non-linear relationships:} Different methods deal with non-linear interactions differently; still, they typically encounter difficulties with intricate and time-varying non-linearities existing in many real systems. Runge et al. (2019) argue that many real systems are characterized by complex, non-linear dynamics that are hard to capture through classical causal discovery procedures. It should be noted that time series data is subject to very high noise levels, and shifts in regime often occur, making it very difficult to find stable causal relations even among the most clever of models.

    \item \textbf{Computational scalability:} As the dimensionality of the system increases, many causal discovery approaches are found to be challenged by a marked deterioration of efficiency. This limitation not only affects the computational tractability of these methods but also their ability to accurately identify causal relationships in high-dimensional data. The experiments, conducted in Chapter 4, confirm that the capability of causal discovery methods weakens as the dimensionality increases with the number of variables. This limitation makes it difficult to apply these methods to extensive time series analyses, especially in areas like finance and climate science with high-dimensional data. While our approach aims to enhance accuracy, we want to strike a balance between improved accuracy and reasonable computational demands.
    
    \item \textbf{Lack of robustness across different data subsets:} Causal relationships inferred from one subset of data may not hold for another, indicating a lack of robustness \cite{Spirtes2016}. This lack of consistency highlights a significant problem in accurately identifying causal relations. It's particularly troublesome in rapidly evolving environments, like those seen in financial markets.
\end{itemize}

\subsection{Other Notable Limitations}

While not the primary focus of our RCV method, the following limitations are also significant in the field of causal discovery:

\begin{itemize}
    \item \textbf{Assumption of causal sufficiency:} Many approaches assume the sufficiency of the causal relation. The assumption is usually untrue regarding complex systems, where hidden variables can play crucial roles \cite{Peters2017, Cai2019}.

    \item \textbf{State-dependent coupling:} Complex systems often exhibit state-dependent relationships between variables. The character of these connections can change over time. In some states, variables might show positive coupling. In others, they may appear unrelated or even exhibit negative coupling, depending on the system's overall condition \cite{Sugihara2012}. This dynamic behaviour creates significant challenges for conventional causal discovery methods that assume stable relationships.
\end{itemize}

\subsection{Implications and Way Forward}
These limitations underscore the need for causal discovery techniques that are both robust and adaptable. Such methods must be capable of handling the complex dynamics found in systems like finance, climate science, and neuroscience. Developing these techniques is important for enhancing our understanding of these systems and improving the validity of complex system models.
\\ \\
In the following sections, we will present our Robust Cross-Validated (RCV) method. This approach directly tackles key challenges: sensitivity to noise, handling of non-linear relationships, robustness across data subsets, and computational scalability. The theoretical foundations of our approach will be discussed in Section 3.2, followed by a detailed description of the RCV method in Section 3.3.

\section{Theoretical Foundations}

The Robust Cross-Validated (RCV) causal discovery method is built upon several key theoretical concepts. These concepts directly shape its design and implementation. This section will outline these fundamental ideas, and show how each element helps tackle the challenges we've identified in causal discovery for complex systems.

\subsection{Potential of Cross-Validation for Causal Discovery}
The core of our RCV method is inspired by the principles of cross-validation. This technique is widely used in machine learning and statistical modelling to evaluate model performance and generalisability \cite{Stone1974}. It involves splitting the data into subsets, using a subset for model training and the rest for validation. This approach offers several benefits:

\begin{itemize}
    \item It helps assess how well a model performs on new, unseen data, which reduces the risk of overfitting \cite{Arlot2010}.
    
    \item It provides a more reliable estimate of model performance compared to using just one train-test split.
    
    \item It allows comparison between different models or model configurations \cite{Kohavi1995}.
    
\end{itemize}

Given these benefits, we believe cross-validation could greatly improve causal discovery methods, especially for time series data. By applying cross-validation to causal discovery:

\begin{itemize}
    \item We could potentially identify causal relationships that remain consistent across different data subsets.
    
    \item It could help reduce the risk of inferring false causal relationships due to noise or peculiarities in a specific dataset.
    
    \item It could provide a measure of confidence in the causal structures discovered.
\end{itemize}

However, using cross-validation for causal discovery in time series data poses unique challenges. Unlike predictive modelling, where the goal is to minimize prediction errors, causal discovery aims to accurately identify the underlying causal structure. This requires careful thought about how to divide time series data and measure the consistency of causal structures across different subsets.

\subsection{Robust Statistics in Causal Inference}
Robust statistics offers a framework for creating methods that are more resilient to deviations from model assumptions and the presence of outliers or noise in data. This approach is especially relevant for causal discovery in complex systems such as financial markets, where data often doesn't conform to ideal conditions \cite{Huber2011}.
\\ \\
Key concepts from robust statistics that can be applied to causal discovery include:

\begin{itemize}
    \item Resistance to outliers: Methods that are less influenced by extreme observations or anomalies in the data. (This concept is incorporated into the RCV method through the Validate\_And\_Adjust process, as detailed in Section 3.3.3)

    \item Robust estimators: Statistical estimators that perform well under a wide range of conditions, not just under ideal circumstances.
    
    \item Distribution-free methods: Techniques that do not rely on strong assumptions about the underlying data distribution.
\end{itemize}

Incorporating these robust statistical principles into causal discovery algorithms can lead to more reliable and stable causal inferences, especially in the context of financial ABMs where data can be noisy and non-stationary \cite{Peters2017}.

\subsection{Causal Consistency and Variability in Time Series}

When analysing time series data from various fields, it's important to consider two key aspects: how consistent causal relations are over time, and how they vary across different data subsets. These concepts are central to our RCV method's approach to robust causal discovery.
\\ \\
Causal consistency measures how stable causal relationships remain over time or across different data regimes. In many real-world time series, consistency challenges arise due to changing conditions, cyclical patterns, and evolving system dynamics. Our RCV method addresses this through the Consistency metric (Equation \ref{eq:consistency} in Section 3.3.2), which measures how often a causal effect's direction is preserved across different data subsets.
\\ \\
Complementary to consistency, causal variability refers to how much inferred causal structures change when the data is slightly altered or when the discovery method is changed. Low variability, or high stability, is crucial for ensuring that discovered causal relations are meaningful and not just artefacts of a specific data sample. The RCV method captures this through the Variability metric (Equation \ref{eq:variability} in Section 3.3.2), which measures the stability of the estimated effect.
\\ \\
To tackle these issues effectively, our method provides measures of uncertainty or confidence in the inferred causal relationships via the consistency and variability metrics. It also allows for the incorporation of domain knowledge about the stability of certain causal mechanisms through adjustable thresholds $\tau_c$ and $\tau_v$ (Algorithm \ref{alg:rcv}, Section 3.3.3).
\\ \\
By focusing on these aspects, our RCV method aims to develop more reliable causal discovery techniques for complex time series data. This leads to improved model validation and a deeper understanding of dynamic systems across various fields. The next section outlines our proposed methodology, which addresses these challenges by incorporating cross-validation and robust statistical techniques into the causal discovery process.

\subsection{Proposed Methodology Overview}
We propose a robust cross-validated approach to causal discovery based on the theoretical foundations discussed in the previous sections. This method integrates the principles of cross-validation with robust statistical techniques to address the challenges of causal discovery in complex time series data.

\begin{figure}[htbp]
\centering
\includegraphics[width=\textwidth]{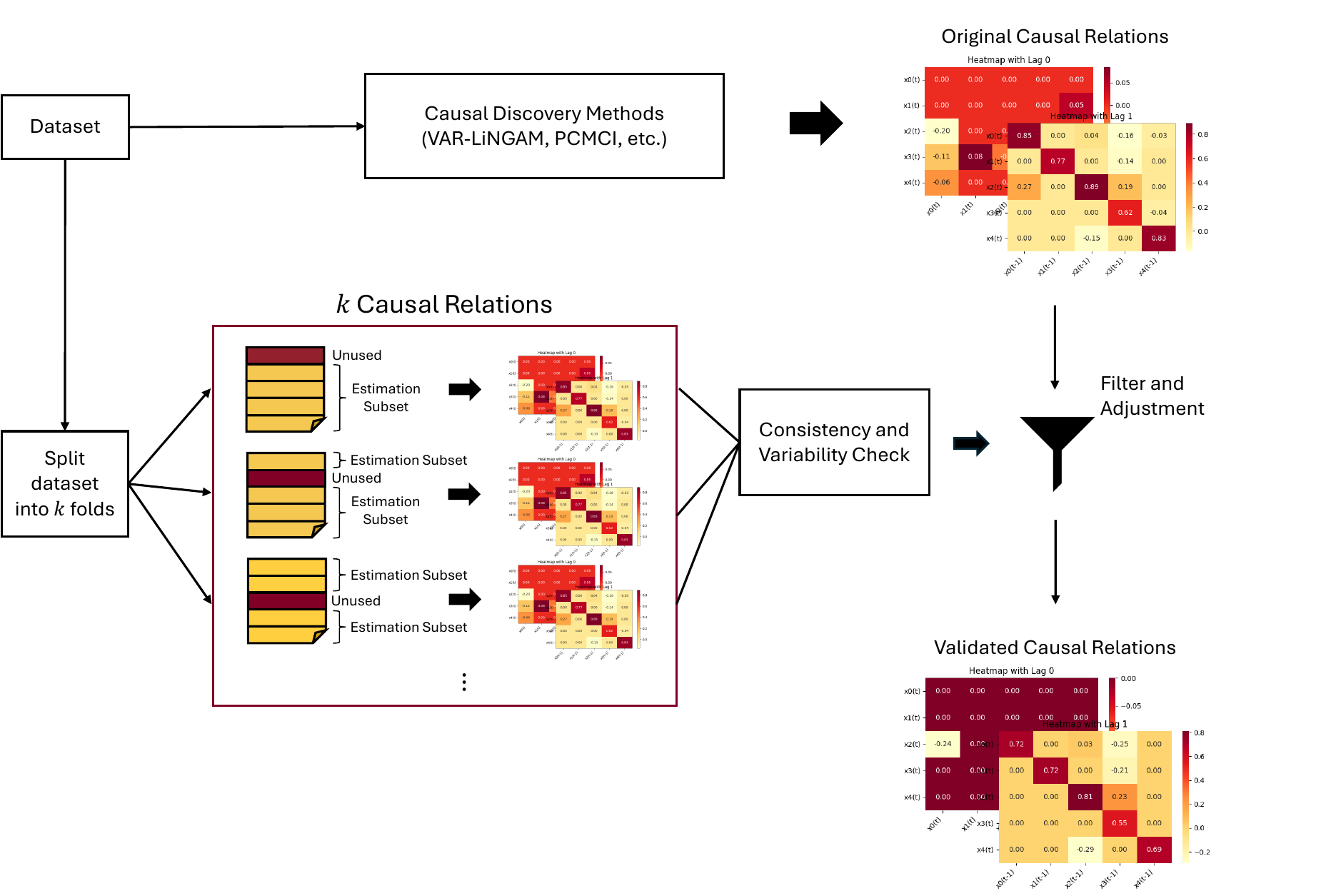}
\caption{Workflow of the Robust Cross-Validated (RCV) causal discovery approach. The process involves initial causal discovery on the full dataset, k-fold cross-validation with independent causal analysis on each fold, consistency and variability checks, and a final filtering and adjustment step to produce validated causal relations.}
\label{fig:rcv_workflow}
\end{figure}

Figure \ref{fig:rcv_workflow} presents an overview of our proposed approach. First, we start with the input dataset. This data is run through a chosen causal discovery method, such as VAR-LiNGAM or PCMCI, to get an initial set of causal relations. At the same time, we split the dataset into k parts for cross-validation. We analyze each part independently using the same causal discovery method. This gives us k sets of causal relations. Next, we take these k sets and the original causal relations and put them through a consistency and variability check. This step helps us see how stable the identified causal relationships are across different subsets of the data. Finally, we apply a filtering and adjustment process. This step takes into account the results from the cross-validation. The outcome is a set of validated causal relations. This approach aims to enhance the robustness and reliability of the discovered causal structures in complex time series data.

\section{The Robust Cross-Validated Method}

\subsection{Overview of the RCV Approach}
The Robust Cross-Validated (RCV) causal discovery method is a novel approach designed to enhance the reliability and stability of causal inference in complex systems, particularly financial Agent-Based Models (ABMs). The RCV method integrates cross-validation techniques with existing causal discovery algorithms to address the challenges of noise sensitivity, overfitting, and lack of robustness often encountered in traditional methods.
\\ \\
Key components of the RCV approach include:
\begin{itemize}
    \item Application of a base causal discovery method to the full dataset.
    \item K-fold cross-validation to assess consistency and variability of causal relationships (refer to Section 3.2.1 for the cross-validation strategy).
    \item A validation and adjustment process to refine the initial causal structure.
\end{itemize}

The RCV method aims to produce causal structures that are more robust to noise, more consistent across different data subsets, and thus more reliable for use in ABM validation and financial system analysis.

\subsection{Mathematical Formulation}

Let $\mathcal{D}$ be the original dataset, and $\mathcal{D}_1, \mathcal{D}_2, ..., \mathcal{D}_k$ be $k$ subsets of $\mathcal{D}$ used for cross-validation. For each potential causal relationship $r_{ij}$ between variables $i$ and $j$, we define two key metrics:

\begin{equation}
    \text{Consistency}(r_{ij}) = \frac{1}{k} \sum_{m=1}^k \mathbf{1}\{\text{sign}(r_{ij}^m) = \text{sign}(r_{ij}^0)\}
    \label{eq:consistency}
\end{equation}

\begin{equation}
    \text{Variability}(r_{ij}) = \frac{\text{std}(\{r_{ij}^m\}_{m=1}^k)}{|r_{ij}^0| + \epsilon}
    \label{eq:variability}
\end{equation}

where:
\begin{itemize}
    \item $r_{ij}^0$ is the initial estimate of the causal effect from the full dataset
    \item $r_{ij}^m$ is the estimate from the $m$-th fold
    \item $\mathbf{1}\{\cdot\}$ is the indicator function
    \item $\text{std}(\cdot)$ denotes the standard deviation
    \item $\epsilon$ is a small constant to avoid division by zero
    \\
\end{itemize}
The Consistency metric measures how often the direction of the causal effect is preserved across different subsets, while the Variability metric quantifies the stability of the effect size estimate.
\\ \\
For example, consider a causal relationship $r_{ij}$ between variables $i$ and $j$:

\begin{itemize}
    \item High consistency (close to 1): The causal direction from $i$ to $j$ is consistent across most or all data subsets, indicating a robust relationship.
    \item Low consistency (close to 0): The causal direction varies significantly across subsets, suggesting an unstable or spurious relationship.
    \item Low variability: The strength of the causal effect is similar across subsets, indicating a stable relationship.
    \item High variability: The strength of the effect varies greatly, suggesting sensitivity to data perturbations.
\end{itemize}
In our RCV method, these values are obtained through the Consistency and Variability metrics (Equations \ref{eq:consistency} and \ref{eq:variability}). The Validate\_And\_Adjust step in Algorithm \ref{alg:rcv} uses these metrics to filter out relationships with low consistency or high variability, ensuring that only robust and stable causal relationships are retained in the final model. More specifically, causal relations with consistency below threshold $\tau_c$ are considered unstable and are filtered out, and those with variability above threshold $\tau_v$ are considered too sensitive to data perturbations and are also filtered out.
\\ \\
This approach allows the RCV method to systematically identify and retain only those causal relationships that demonstrate both directional consistency and strength stability across different subsets of the data.

\subsection{Algorithm Description}

The RCV algorithm can be described as follows:

\begin{algorithm}
\caption{Robust Cross-Validated Causal Discovery}
\label{alg:rcv}
\begin{algorithmic}[1]
\Require Dataset $\mathcal{D}$, number of folds $k$, consistency threshold $\tau_c$, variability threshold $\tau_v$
\Ensure Robust causal structure $G$
\State $G_0 \gets \text{Initial\_Causal\_Discovery}(\mathcal{D})$
\For{$i = 1$ to $k$}
    \State $\mathcal{D}_i \gets \text{Split\_Dataset}(\mathcal{D}, i)$
    \State $G_i \gets \text{Causal\_Discovery}(\mathcal{D}_i)$
\EndFor
\State $G \gets \text{Validate\_And\_Adjust}(G_0, \{G_i\}_{i=1}^k, \tau_c, \tau_v)$
\State \Return $G$
\end{algorithmic}
\end{algorithm}

The `Validate\_And\_Adjust' function is a critical component that applies the consistency and variability checks to each causal relationship, retaining only those that meet the specified thresholds. It may also adjust the strength of retained relationships based on the cross-validation results.
\\ \\
The benefits and effectiveness of the RCV method, including its robustness to noise, ability to handle non-linear relationships, and improved computational scalability, are experimentally realised and quantified in Chapter 4's comprehensive evaluations.

\subsection{Implementation Details}

The RCV method has been implemented with two prominent causal discovery algorithms: VAR-LiNGAM and PCMCI. This dual implementation demonstrates the flexibility of the RCV approach and allows for a comparative analysis of its effectiveness across different base methods. A comprehensive evaluation of these implementations, including a detailed comparison with their original counterparts, is presented in Chapter 4.

\subsubsection{Base Causal Discovery Methods}

\begin{enumerate}
    \item \textbf{VAR-LiNGAM:} Vector Autoregressive Linear Non-Gaussian Acyclic Model (VAR-LiNGAM) is chosen for its ability to handle both contemporaneous and time-lagged causal relationships in time series data, making it particularly suitable for financial ABMs \cite{Hyvarinen2010}.

    \item \textbf{PCMCI:} The Peter and Clark algorithm with Momentary Conditional Independence (PCMCI) is selected for its effectiveness in dealing with high-dimensional time series data and its ability to control for autocorrelation \cite{Runge2019}.
\end{enumerate}

The application of RCV to both methods allows us to assess its performance across different underlying assumptions and methodologies.

\subsubsection{Cross-Validation Strategy}
For both VAR-LiNGAM and PCMCI implementations, we employ a standard k-fold cross-validation approach. Adopting this approach allows us to assess the consistency and stability of causal relationships across different subsets of the data, with the overall structure of the time series still preserved.
\\ \\
The number of folds (k) in the cross-validation process is a critical parameter. The procedure of tuning on synthetic data empirically has shown that, for time series data of length 1000, k = 7 was able to balance out the optimal results. Nevertheless, this optimal k should also take into account the specific nature of the data and the causal discovery algorithm used. The choice of k involves a trade-off between computational cost and the stability of the cross-validation results.

\subsubsection{Threshold Selection}
The choice of consistency ($\tau_c$) and variability ($\tau_v$) thresholds is crucial and can be determined through hyperparameter tuning:

\begin{itemize}
    \item For VAR-LiNGAM: We found optimal values of $\tau_c = 0.4$ and $\tau_v = 0.4$
    
    \item For PCMCI: We found optimal values of $\tau_c = 0.7$ and $\tau_v = 0.4$
\end{itemize}

The lower consistency threshold for VAR-LiNGAM compared to PCMCI aligns with observations in the causal discovery literature. As noted by \citet{Assaad2022}, ``some methods are more precision-oriented, as they detect few but relevant relations (TCDF, VarLiNGAM), whereas other methods are more recall-oriented and focus on the detection of all relevant relations." Given that VAR-LiNGAM tends to identify fewer but more relevant causal relationships, a lower consistency threshold is sufficient to ensure the reliability of the discovered causal structure.

\subsubsection{Adjustment Process}
The adjustment process involves a weighted average of the initial estimate and the mean of the cross-validation estimates:

\begin{equation}
    r_{ij}^{adjusted} = (1-w) \cdot r_{ij}^0 + w \cdot \text{mean}(\{r_{ij}^m\}_{m=1}^k)
\end{equation}

where $w$ is a weighting factor. A larger $w$ gives more weight to the results from the cross-validation subsets, while a smaller $w$ favours the initial estimate from the full dataset. In this study, as we focus primarily on the structure of causal relationships rather than their magnitudes, we set $w = 0$, effectively using only the initial estimate:

\begin{equation}
    r_{ij}^{adjusted} = r_{ij}^0
\end{equation}

This choice aligns with our research objectives of identifying robust causal structures. Since RCV method is flexible, the designer would be able to employ the effect size adjustments in the future. In contexts where effect sizes are of interest, typical values for $w$ might range from 0.05 to 0.5, balancing the initial estimate with cross-validated results. We may use higher values of $w$ when there is a need for greater precision to account for the variations across different subsets of the data.
\\ \\
This dual implementation approach enhances the versatility and applicability of the RCV method in the context of ABM validation and causal discovery in complex systems. The application of the new technique is not limited to the methods of RCV but can be utilized in the other base methods of causal discovery.

\section{Potential Impact on ABM Validation}

The Robust Cross-Validated (RCV) method has the potential to significantly enhance the validation process of Agent-Based Models (ABMs) across various domains, including but not limited to finance:

\subsection{Improved Causal Structure Identification}

RCV offers several advantages in identifying causal structures:

\begin{itemize}
    \item \textbf{Robustness to Noise:} By leveraging cross-validation, RCV can better distinguish genuine causal relationships from spurious correlations induced by noise.
    
    \item \textbf{Consistency Across Data Subsets:} RCV's ability to assess consistency across different data subsets can help identify stable causal relationships that persist across various conditions.
    
    \item \textbf{Uncertainty Quantification:} The consistency and variability metrics provided by RCV offer a measure of confidence in each inferred causal link.
\end{itemize}

\subsection{Enhanced Model Comparison and Validation}

RCV provides a framework for more rigorous ABM comparison and validation:

\begin{itemize}
    \item \textbf{Standardized Metrics:} The consistency and variability scores from RCV can serve as standardized metrics for comparing causal structures inferred by different models.
    
    \item \textbf{Cross-Model Validation:} Causal structures inferred from one model can be tested for consistency in data generated by another model or in empirical data.
    
    \item \textbf{Iterative Refinement:} Detailed feedback on the stability of each causal link can guide focused improvements in model development.
\end{itemize}

\subsection{Addressing Complex System Challenges}

RCV is well-suited to address challenges common in complex systems modelling:

\begin{itemize}
    \item \textbf{Non-stationarity:} RCV can potentially capture time-varying causal relationships, providing a more realistic representation of dynamic systems.
    
    \item \textbf{Heterogeneous Components:} RCV can help identify robust causal patterns that emerge from the collective behaviour of diverse components.
    
    \item \textbf{Complex Interdependencies:} RCV's compatibility with methods like VAR-LiNGAM allows for identifying both contemporaneous and lagged causal relationships, which is crucial in understanding complex system dynamics.

\end{itemize}

By addressing these challenges, RCV can contribute to the development of more accurate and reliable ABMs across various fields, potentially improving our understanding of complex system dynamics and enhancing the tools available for policy analysis and decision-making.

\chapter{Experimental Evaluation}

This chapter details the second major contribution of our research: a Comprehensive Experimental Evaluation and Analysis of causal discovery methods. We present a thorough empirical analysis of our proposed RCV methods and existing approaches across a wide range of dataset characteristics, including linearity, noise distribution, stationarity, and causal structure density. This analysis extends to both synthetic datasets and a complex simulated fMRI dataset, providing insights into method performance under various conditions and addressing the challenge of understanding how dataset properties affect causal discovery performance in ABM validation.

\section{Experimental Setup}

\subsection{Synthetic Dataset Generation}

We developed a comprehensive synthetic dataset generator to evaluate the performance of various causal discovery methods thoroughly. This generator allows us to create time series data with controllable characteristics, enabling systematic evaluation across various scenarios. The use of synthetic data is crucial in our study, as the dynamic interactions in real-world data often do not guarantee a well-defined ground truth.

\subsubsection{Data Generation Process}

The core of our data generation process is based on a Vector Autoregressive (VAR) model with the following key features:

\begin{itemize}
    \item \textbf{Flexible Structure}: The generator allows for multiple lags and instantaneous effects, represented by a series of coefficient matrices $B_0, B_1, ..., B_p$, where $p$ is the maximum lag.
    
    \item \textbf{Instantaneous Effects}: The $B_0$ matrix, representing instantaneous causal effects, is constrained to be lower triangular to ensure identifiability.
    
    \item \textbf{Stability}: We implement a stabilization procedure to ensure that the generated VAR process is stationary, preventing divergence over time. The coefficient matrices are adjusted if necessary to maintain this stability.
    
    \item \textbf{Non-linear Relations}: Optional non-linear effects can be introduced using a hyperbolic tangent function, allowing us to test the methods' performance on more complex relationships.
    
    \item \textbf{Non-Gaussian Noise}: The generator can produce both Gaussian and non-Gaussian (t-distributed) noise, mimicking the diverse noise distributions found in real-world data.
    
    \item \textbf{Non-stationarity}: Trends can be added to simulate non-stationary behaviour, a common characteristic in many real-world time series.
    
    \item \textbf{Fluctuations}: Fluctuations are applied to all variables to mimic random walk behaviour, better demonstrating the variability and unpredictability often present in real-world datasets.
\end{itemize}

This comprehensive set of features allows us to generate synthetic datasets that closely resemble the complexity and variability of real-world time series while maintaining control over the underlying causal structure. This control is essential for accurately evaluating and comparing the performance of different causal discovery methods.

\subsubsection{Controllable Parameters}

Our generator incorporates a range of adjustable parameters, enabling the creation of diverse datasets that simulate various real-world scenarios:

\begin{itemize}
    \item \textbf{Linearity}: Toggle between linear and non-linear relationships, allowing us to evaluate method performance on both simple and complex causal structures.
    
    \item \textbf{Noise Distribution}: Choose between Gaussian and non-Gaussian (t-distributed) noise to mimic different types of random fluctuations observed in real-world data.
    
    \item \textbf{Stationarity}: Option to include trends, enabling the generation of non-stationary data. This feature is crucial for testing methods' ability to handle time-varying causal relationships.
    
    \item \textbf{Causal Density}: Control the sparsity of causal connections to create scenarios ranging from simple, sparse causal networks to complex, densely connected systems.
    
    \item \textbf{Number of Variables}: To assess scalability, we generate datasets with varying numbers of variables. Specifically, we create datasets with 5, 20, and 50 variables, allowing us to evaluate how different methods perform as the problem complexity increases.
    
    \item \textbf{Time Series Length}: We can adjust the temporal extent of the generated datasets. In our experiments, we create time series with lengths of 250, 1000, and 2000 time points, enabling us to investigate the impact of data volume on causal discovery performance.
\end{itemize}

\subsubsection{Dataset Types}

We generated the following types of synthetic datasets:

\begin{enumerate}
    \item \textbf{Linear vs Non-linear}: Testing methods' ability to handle different functional relationships.
    \item \textbf{Gaussian vs Non-Gaussian Noise}: Evaluating performance under different noise assumptions.
    \item \textbf{Stationary vs Non-stationary}: Assessing robustness to time-varying processes.
    \item \textbf{Sparse vs Dense Causal Structures}: Examining performance on different network densities.
    \item \textbf{Variable Scale}: Datasets with 5, 20, and 50 variables to test scalability.
    \item \textbf{Time Series Length}: Datasets with 250, 1000, and 2000 time points to assess the impact of data volume.
\end{enumerate}

The choice of 5, 20, and 50 variables as well as the selected time series lengths of 250, 1000, and 2000 points for our scalability analysis are grounded in common practices in causal discovery literature and reflect realistic scenarios in various domains. Smaller networks with 5 variables are often used in benchmark studies, while networks of 20-50 variables are common in many real-world applications \cite{Shimizu2006}. As for the selected time series lengths, these choices align with various practical scenarios: 250 points might represent a year of daily financial data, 1000 points could correspond to about four years of daily data or a longer period of weekly data, while 2000 points could represent long-term studies or high-frequency data. Similar ranges have been used in other causal discovery studies \cite{Guerini2017, Runge2019}. These ranges allow us to assess method performance from simple to more complex, high-dimensional scenarios.
\\ \\
For each configuration, we generated 10 distinct datasets to ensure robust evaluation and account for random variations in the data generation process.

\subsubsection{Ground Truth}

The true causal structure, represented by coefficient matrices ($n$ matrices represent 1 instantaneous relations and $n-1$ lagged relations), was preserved for each dataset type. This ground truth is crucial for accurately evaluating the performance of causal discovery methods, providing a benchmark against which the inferred causal relationships can be compared.
\\ \\
This comprehensive framework enables systematic evaluation of causal discovery methods across a wide range of controlled scenarios, offering insights into their performance, limitations, and scalability under various data conditions.

\subsection{Simulated fMRI Dataset Description}

To complement our synthetic data experiments, we utilized a simulated fMRI (Functional Magnetic Resonance Imaging) dataset, which provides a valuable test case for our methods due to its complex temporal dynamics and known causal relationships between simulated brain regions. This dataset, originally developed by Smith et al. \cite{Smith2011} and preprocessed by Nauta et al. \cite{Nauta2019}, contains simulated BOLD (Blood-oxygen-level dependent) data for 27 different underlying brain networks.
\\ \\
Each dataset represents simulated neural activity based on modelled blood flow changes in different brain regions. The number of variables (time series) ranges from 5 to 15, with each variable corresponding to a distinct simulated brain region. The time series lengths vary between 50 and 5000 time points, with a mean of 774. The datasets exhibit non-linear relationships and include both self-causation and confounders, making them particularly challenging and realistic for causal discovery evaluation.
\\ \\
Importantly, these simulated fMRI datasets are causally sufficient, meaning that all common causes are included in the simulation. The number of causal relationships in each dataset varies, ranging from 10 to 33. Due to computational constraints of some methods, we excluded larger datasets, focusing on those with at most 15 time series \cite{Nauta2019}.
\\ \\
This simulated benchmark enables us to evaluate the performance of causal discovery methods on data characterized by complex, non-linear interactions and diverse temporal dynamics while providing ground truth causal structures for accurate performance assessment. It offers valuable insights into the applicability of these methods in scenarios that closely mimic the complexities of real-world fMRI data.

\subsection{Performance Metrics}

To comprehensively evaluate the performance of the causal discovery methods, we employed a set of complementary metrics:

\begin{itemize}
    \item \textbf{Structural Hamming Distance (SHD)}: Quantifies the total number of edge additions, deletions, and reversals needed to transform the estimated graph into the true graph. It offers a holistic measure of structural differences.

    \item \textbf{F1-score}: Computed based on the presence or absence of edges, regardless of direction. This metric provides insight into the method's ability to detect the existence of relationships, even if the direction is the opposite.
    
    \item \textbf{F1-score Directed (F1\_directed)}: Our primary focus metric, calculated using the sign of the causal effect matrices. It measures the accuracy of both the presence and direction of causal relationships, providing a balanced view of precision and recall in the directed graph structure.

    \item \textbf{Frobenius Norm}: While not included in our main analysis tables, this metric is calculated to measure the overall difference in magnitude between the true and estimated causal effect matrices.

    \item \textbf{Runtime}: Measures the computational efficiency of each method, which is crucial for assessing scalability and practical applicability.
\end{itemize}

Our selection of performance metrics is based on their widespread use and effectiveness in evaluating causal discovery methods. The Structural Hamming Distance (SHD) is a standard way to compare graph structures, and it has been widely used in causal discovery literature \citep{Tsamardinos2006}. F1 scores, both directed and undirected, give us a balanced view of precision and recall. This is crucial for judging how accurately a method identifies causal relationships. We also included runtime as a metric, this allows us to evaluate computational efficiency, which is an important factor in large-scale analyses. Together, these metrics offer a comprehensive evaluation of both accuracy and efficiency. They enable a thorough comparison of different causal discovery methods.
\\ \\
To illustrate the difference between the standard F1 score and the F1-directed score, let's consider the following example:
\\ \\
Suppose the true causal relationships are:
\begin{equation}
A \rightarrow B, \quad B \rightarrow C, \quad C \rightarrow D
\end{equation}

Our method infers the following causal relationships:
\begin{equation}
A \rightarrow B, \quad B \rightarrow C, \quad D \rightarrow C
\end{equation}

For the standard F1 score calculation:
\begin{itemize}
\item True Positives (TP) = 3 (A-B, B-C, and C-D relationships are identified, regardless of direction)
\item False Positives (FP) = 0 (no extra relationships are inferred)
\item False Negatives (FN) = 0 (no relationships are missed)
\end{itemize}

Therefore:
\begin{equation}
\text{Precision} = \frac{\text{TP}}{\text{TP} + \text{FP}} = \frac{3}{3 + 0} = 1
\end{equation}

\begin{equation}
\text{Recall} = \frac{\text{TP}}{\text{TP} + \text{FN}} = \frac{3}{3 + 0} = 1
\end{equation}

The standard F1 score is calculated as:
\begin{equation}
\text{F1-standard} = 2 \cdot \frac{\text{Precision} \cdot \text{Recall}}{\text{Precision} + \text{Recall}} = 2 \cdot \frac{1 \cdot 1}{1 + 1} = 1
\end{equation}

For the F1-directed score calculation:
\begin{itemize}
\item True Positives (TP) = 2 ($A \rightarrow B$ and $B \rightarrow C$ are correctly identified)
\item False Positives (FP) = 1 ($D \rightarrow C$ is incorrectly identified)
\item False Negatives (FN) = 1 ($C \rightarrow D$ is not identified)
\end{itemize}

Therefore:
\begin{equation}
\text{Precision} = \frac{\text{TP}}{\text{TP} + \text{FP}} = \frac{2}{2 + 1} = \frac{2}{3}
\end{equation}

\begin{equation}
\text{Recall} = \frac{\text{TP}}{\text{TP} + \text{FN}} = \frac{2}{2 + 1} = \frac{2}{3}
\end{equation}

The F1-directed score is calculated as:
\begin{equation}
\text{F1-directed} = 2 \cdot \frac{\text{Precision} \cdot \text{Recall}}{\text{Precision} + \text{Recall}} = 2 \cdot \frac{\frac{2}{3} \cdot \frac{2}{3}}{\frac{2}{3} + \frac{2}{3}} = \frac{2}{3} \approx 0.667
\end{equation}

This example illustrates a crucial difference between the two metrics. The standard F1 score yields a perfect score of 1, suggesting flawless performance, as it only considers the presence of relationships between variables, regardless of their direction. In contrast, the F1-directed score of approximately 0.667 reflects the method's imperfect performance in identifying the correct causal directions.
\\ \\
The F1-directed score penalizes both incorrect directions (D → C instead of C → D) and missed relationships, making it more suitable for evaluating causal discovery methods where the direction of relationships is crucial. The standard F1 score, while useful in many contexts, may overestimate performance in causal discovery tasks by ignoring directional errors.
\\ \\
Researchers should choose the appropriate metric based on their specific requirements and the importance of directional accuracy in their causal models. In the context of causal discovery, particularly for complex systems like those found in economics and finance, the F1-directed score provides a more stringent and relevant evaluation of model performance.
\\ \\

\subsection{Explanation of result production}
Our results were produced through a rigorous experimental process. For each dataset type and scale, we generated 10 distinct synthetic datasets using the process described in Section 4.1.1. Each causal discovery method was then applied to these datasets. The performance metrics (SHD, F1 scores, runtime) were calculated for each run, and the mean and standard deviation across the 10 runs were computed. This approach ensures robustness against random variations in data generation and provides a measure of the methods' consistency across similar datasets.
\\ \\
For the different dataset characteristics (linear vs non-linear, Gaussian vs non-Gaussian noise, etc.), we used the same data generation process but modified the relevant parameters. For instance, non-linear relationships were introduced using a hyperbolic tangent function applied to the linear combinations, while non-Gaussian noise was generated using a t-distribution instead of a Gaussian distribution.
\\ \\
All experiments were conducted on a MacBook Pro 14-inch 2023 model, equipped with an Apple M2 Pro chip and 32GB of RAM, running macOS Ventura. The algorithms were implemented in Python 3.8 and executed within a virtual environment (.venv) in Visual Studio Code. Key Python libraries utilized include numpy (1.24.4), pandas (2.0.3), and scikit-learn (1.3.2).
\\ \\
For the implementation of specific causal discovery methods, we utilized the following packages:

\begin{itemize}
    \item VARLiNGAM: We used the LiNGAM (Linear Non-Gaussian Acyclic Model) package (version 1.5.2) available at \url{https://github.com/cdt15/lingam}. This package provides efficient implementations of various LiNGAM algorithms, including VARLiNGAM.

    \item PCMCI: We employed the Tigramite package (version 4.2.0), accessible at \url{https://github.com/jakobrunge/tigramite}. Tigramite is a causal time series analysis python package, which includes the implementation of the PCMCI algorithm.
\end{itemize}

These packages ensured that we used state-of-the-art implementations of VARLiNGAM and PCMCI, contributing to the reliability and reproducibility of our results. For our proposed RCV methods, we extended these base implementations, incorporating our cross-validation approach as described in Section 3.

\section{Comparative Analysis}
In this section, we present a comprehensive comparison of five causal discovery methods: VAR-LiNGAM, VAR-LiNGAM Bootstrap, RCV-VAR-LiNGAM (our proposed method), PCMCI, and RCV-PCMCI (our proposed method).
\\ \\
We evaluate these methods across different dataset characteristics and scales to assess their strengths and limitations.

\subsection{Performance Across Different Dataset Characteristics}

\subsubsection{Linear vs Non-linear Relationships}
The results in Table \ref{tab:linear_nonlinear_performance} show that RCV-VAR-LiNGAM consistently outperforms other methods in both linear and non-linear scenarios.

\begin{table}[htbp]
\centering
\small
\begin{tabular}{llccc}
\toprule
Characteristic & Method & SHD & F1 Score & F1\_directed \\
\midrule
\multirow{5}{*}{Linear} 
 & VAR-LiNGAM & 6.80 $\pm$ 3.74 & 0.771 $\pm$ 0.093 & 0.758 $\pm$ 0.109 \\
 & VL-Bootstrap & 8.50 $\pm$ 2.92 & 0.706 $\pm$ 0.081 & 0.698 $\pm$ 0.081 \\
 & RCV-VAR-LiNGAM & \textbf{3.40 $\pm$ 1.71} & \textbf{0.848 $\pm$ 0.078} & \textbf{0.848 $\pm$ 0.078} \\
 & PCMCI & 13.70 $\pm$ 4.60 & 0.609 $\pm$ 0.082 & 0.591 $\pm$ 0.076 \\
 & RCV-PCMCI & 7.50 $\pm$ 0.85 & 0.695 $\pm$ 0.056 & 0.687 $\pm$ 0.041 \\
\midrule
\multirow{5}{*}{Non-linear} 
 & VAR-LiNGAM & 8.60 $\pm$ 2.32 & 0.719 $\pm$ 0.053 & 0.712 $\pm$ 0.065 \\
 & VL-Bootstrap & 8.10 $\pm$ 3.18 & 0.727 $\pm$ 0.097 & 0.720 $\pm$ 0.101 \\
 & RCV-VAR-LiNGAM & \textbf{4.40 $\pm$ 1.51} & \textbf{0.819 $\pm$ 0.052} & \textbf{0.819 $\pm$ 0.052} \\
 & PCMCI & 14.70 $\pm$ 3.06 & 0.589 $\pm$ 0.079 & 0.574 $\pm$ 0.061 \\
 & RCV-PCMCI & 7.90 $\pm$ 1.45 & 0.665 $\pm$ 0.057 & 0.665 $\pm$ 0.057 \\
\bottomrule
\end{tabular}
\caption{Performance comparison for Linear vs Non-linear datasets}
\label{tab:linear_nonlinear_performance}
\end{table}

\begin{figure}[htbp]
\centering
\begin{subfigure}[b]{0.48\textwidth}
\centering
\includegraphics[width=\textwidth]{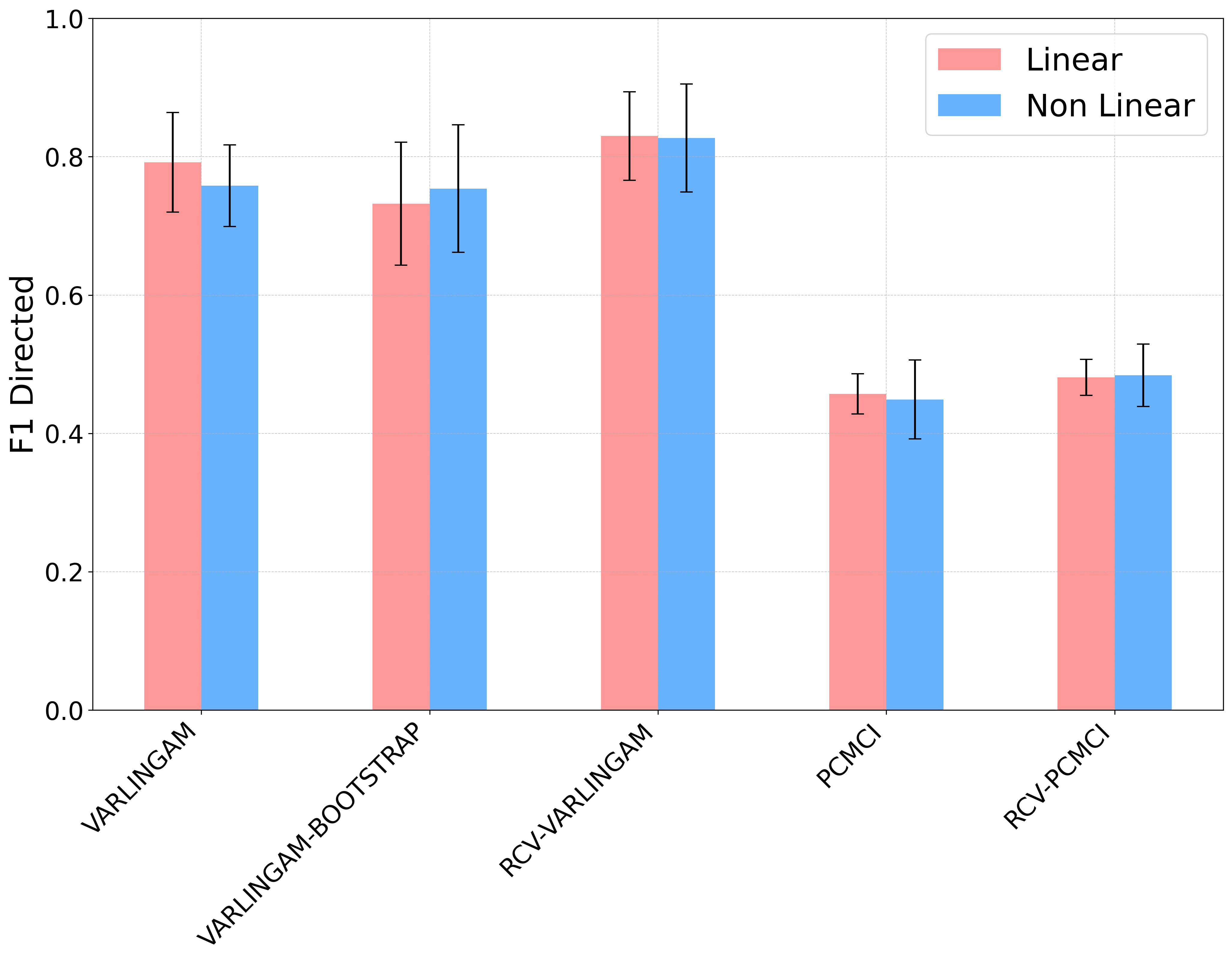}
\caption{F1 Directed Scores}
\label{fig:f1_directed_linear_nonlinear}
\end{subfigure}
\hfill
\begin{subfigure}[b]{0.48\textwidth}
\centering
\includegraphics[width=\textwidth]{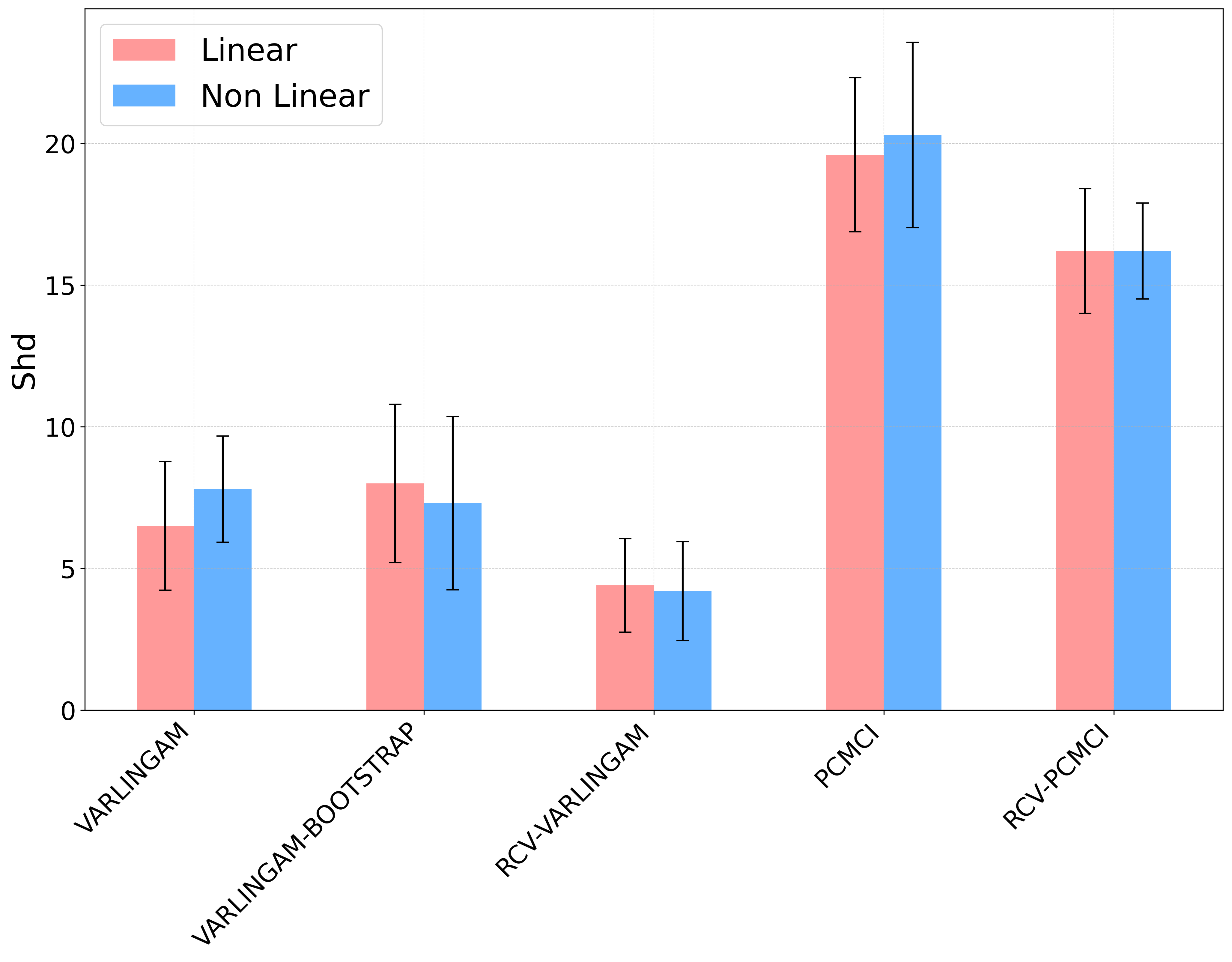}
\caption{Structural Hamming Distance}
\label{fig:shd_linear_nonlinear}
\end{subfigure}
\caption{Comparison of performance metrics for linear and non-linear datasets. (a) F1 Directed scores show the accuracy of causal discovery. (b) SHD values indicate the overall structural difference from the true causal graph, where lower values are better.}
\label{fig:linear_nonlinear_comparison}
\end{figure}

\begin{itemize}
    \item \textbf{RCV-VAR-LiNGAM Superiority}: This method consistently outperforms others, achieving the highest F1 directed scores (0.848 for linear and 0.819 for non-linear) and the lowest SHD values (3.40 and 4.40 respectively). This indicates robust performance in capturing both linear and non-linear causal relations.

    \item \textbf{Resilience to Non-linearity}: While most methods' performance deteriorates in non-linear settings, RCV-VAR-LiNGAM exhibits the smallest decline (F1 directed from 0.848 to 0.819), showing strong adaptability to complex relations.

    \item \textbf{PCMCI Limitations}: Both PCMCI and RCV-PCMCI struggle comparatively, particularly in non-linear scenarios. This suggests these methods may be less suitable for complex, non-linear time series.
\end{itemize}

Figure \ref{fig:linear_nonlinear_comparison} visually reinforces these findings, clearly illustrating RCV-VAR-LiNGAM's dominance in both F1 directed scores and SHD values across linear and non-linear datasets.

\subsubsection{Gaussian vs Non-Gaussian Noise}
The results in Table \ref{tab:gaussian_nongaussian_performance} reveal significant insights into the performance of causal discovery methods under different noise distributions.

\begin{table}[htbp]
\centering
\small
\begin{tabular}{llccc}
\toprule
Characteristic & Method & SHD & F1 Score & F1\_directed \\
\midrule
\multirow{5}{*}{Gaussian} 
 & VAR-LiNGAM & 9.40 $\pm$ 3.66 & 0.686 $\pm$ 0.100 & 0.673 $\pm$ 0.106 \\
 & VL-Bootstrap & 9.20 $\pm$ 2.86 & 0.677 $\pm$ 0.089 & 0.663 $\pm$ 0.089 \\
 & RCV-VAR-LiNGAM & \textbf{5.20 $\pm$ 2.20} & \textbf{0.761 $\pm$ 0.075} & \textbf{0.752 $\pm$ 0.090} \\
 & PCMCI & 13.40 $\pm$ 3.63 & 0.628 $\pm$ 0.067 & 0.599 $\pm$ 0.070 \\
 & RCV-PCMCI & 7.50 $\pm$ 1.08 & 0.672 $\pm$ 0.037 & 0.664 $\pm$ 0.041 \\
\midrule
\multirow{5}{*}{Non-Gaussian} 
 & VAR-LiNGAM & 6.50 $\pm$ 2.01 & 0.758 $\pm$ 0.061 & 0.758 $\pm$ 0.061 \\
 & VL-Bootstrap & 6.60 $\pm$ 2.32 & 0.759 $\pm$ 0.074 & 0.752 $\pm$ 0.073 \\
 & RCV-VAR-LiNGAM & \textbf{4.00 $\pm$ 0.67} & \textbf{0.818 $\pm$ 0.039} & \textbf{0.818 $\pm$ 0.039} \\
 & PCMCI & 14.60 $\pm$ 3.44 & 0.593 $\pm$ 0.079 & 0.573 $\pm$ 0.056 \\
 & RCV-PCMCI & 7.30 $\pm$ 1.49 & 0.699 $\pm$ 0.040 & 0.699 $\pm$ 0.040 \\
\bottomrule
\end{tabular}
\caption{Performance comparison for Gaussian vs Non-Gaussian datasets}
\label{tab:gaussian_nongaussian_performance}
\end{table}

\begin{figure}[htbp]
\centering
\begin{subfigure}[b]{0.48\textwidth}
\centering
\includegraphics[width=\textwidth]{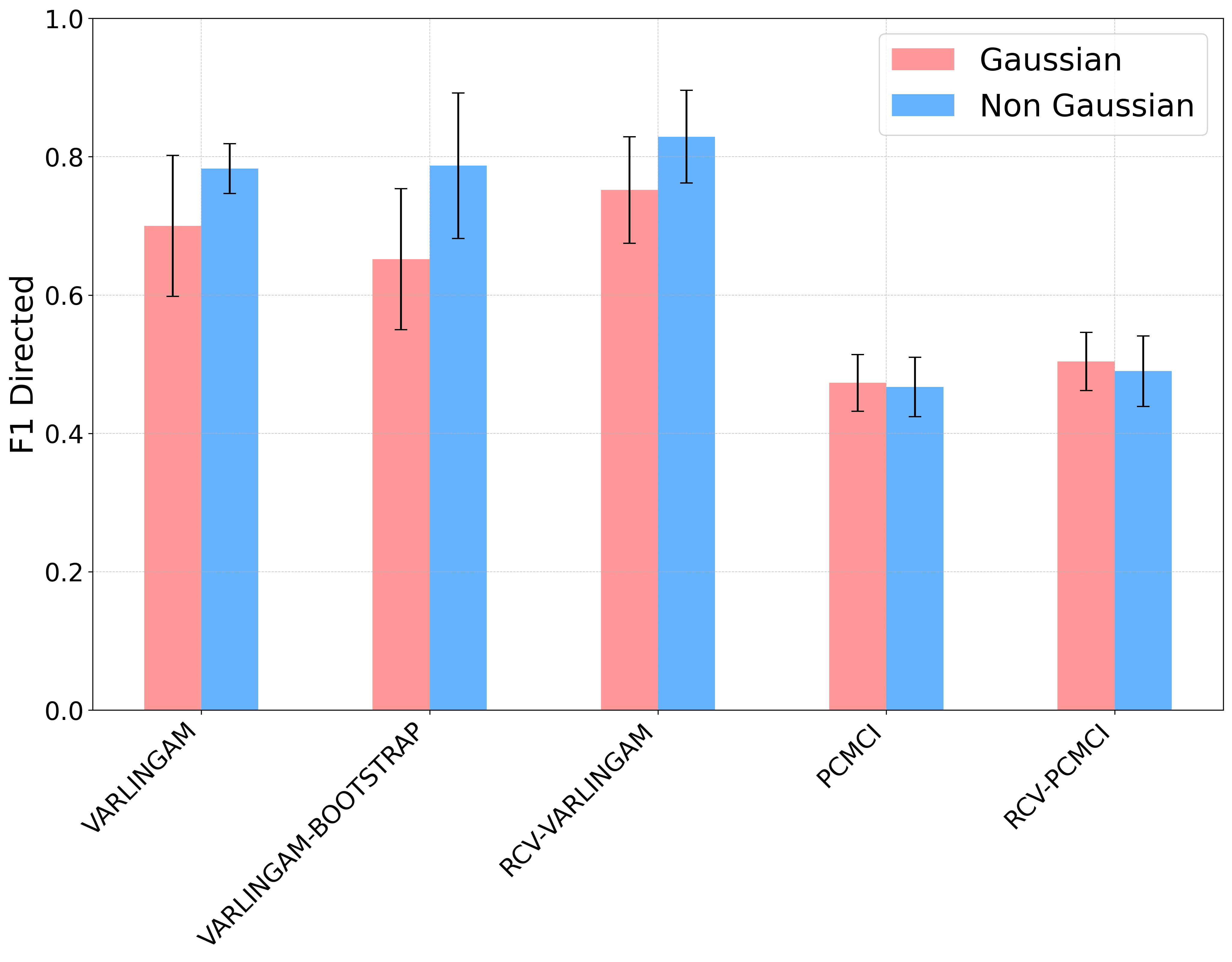}
\caption{F1 Directed Scores}
\label{fig:f1_directed_gaussian_nongaussian}
\end{subfigure}
\hfill
\begin{subfigure}[b]{0.48\textwidth}
\centering
\includegraphics[width=\textwidth]{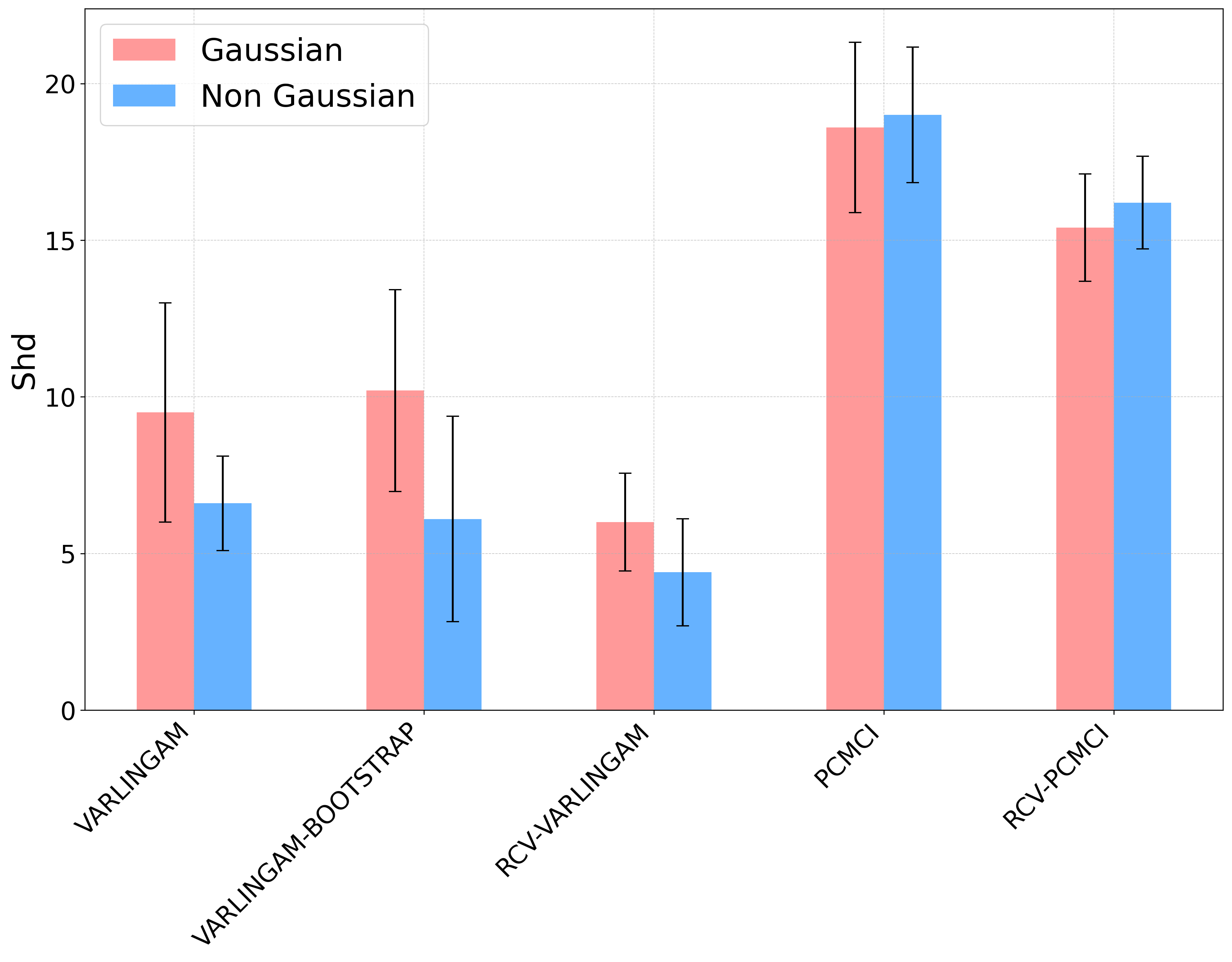}
\caption{Structural Hamming Distance}
\label{fig:shd_gaussian_nongaussian}
\end{subfigure}
\caption{Comparison of performance metrics for Gaussian and non-Gaussian noise datasets. (a) F1 Directed scores show the accuracy of causal discovery under different noise conditions. (b) SHD values indicate the overall structural difference from the true causal graph.}
\label{fig:gaussian_nongaussian_comparison}
\end{figure}

\begin{itemize}
    \item \textbf{Non-Gaussian Advantage}: All methods exhibit improved performance with non-Gaussian noise, aligning with theoretical expectations. This improvement is particularly pronounced for VAR-LiNGAM and its variants, underscoring the methods' reliance on non-Gaussianity for causal inference.

    \item \textbf{RCV-VAR-LiNGAM Robustness}: This method demonstrates superior performance across both noise types, achieving the highest F1 directed scores (0.752 for Gaussian and 0.818 for non-Gaussian). Its consistent performance, indicated by low SHD values (5.20 and 4.00), suggests a robust ability to handle various noise distributions.

    \item \textbf{PCMCI Sensitivity}: Both PCMCI and RCV-PCMCI show higher sensitivity to noise distribution. PCMCI's performance notably degrades in non-Gaussian scenarios (F1 directed from 0.599 to 0.573), contrary to the trend observed in other methods. This suggests potential limitations in PCMCI's ability to leverage non-Gaussian information.
\end{itemize}

Figure \ref{fig:gaussian_nongaussian_comparison} visually reinforces these observations, clearly illustrating the superior performance of RCV-VAR-LiNGAM, particularly in non-Gaussian settings. The consistent improvement across all VAR-LiNGAM variants in non-Gaussian scenarios underscores the importance of considering noise distribution in causal discovery tasks.

\subsubsection{Stationary vs Non-stationary Time Series}
Table \ref{tab:stationary_nonstationary_performance} presents a comparative analysis of causal discovery methods on stationary and non-stationary time series, revealing several key insights:

\begin{table}[htbp]
\centering
\small
\begin{tabular}{llccc}
\toprule
Characteristic & Method & SHD & F1 Score & F1\_directed \\
\midrule
\multirow{5}{*}{Stationary} 
 & VAR-LiNGAM & 6.70 $\pm$ 1.64 & 0.841 $\pm$ 0.035 & 0.841 $\pm$ 0.035 \\
 & VL-Bootstrap & 6.80 $\pm$ 2.44 & 0.835 $\pm$ 0.057 & 0.835 $\pm$ 0.057 \\
 & RCV-VAR-LiNGAM & \textbf{5.00 $\pm$ 1.05} & \textbf{0.864 $\pm$ 0.029} & \textbf{0.864 $\pm$ 0.029} \\
 & PCMCI & 29.30 $\pm$ 2.26 & 0.583 $\pm$ 0.023 & 0.416 $\pm$ 0.022 \\
 & RCV-PCMCI & 25.70 $\pm$ 1.89 & 0.609 $\pm$ 0.020 & 0.433 $\pm$ 0.024 \\
\midrule
\multirow{5}{*}{Non-stationary} 
 & VAR-LiNGAM & 8.30 $\pm$ 3.09 & 0.778 $\pm$ 0.070 & 0.778 $\pm$ 0.070 \\
 & VL-Bootstrap & 8.10 $\pm$ 2.38 & 0.772 $\pm$ 0.057 & 0.772 $\pm$ 0.057 \\
 & RCV-VAR-LiNGAM & \textbf{5.10 $\pm$ 1.66} & \textbf{0.838 $\pm$ 0.054} & \textbf{0.838 $\pm$ 0.054} \\
 & PCMCI & 25.90 $\pm$ 3.96 & 0.602 $\pm$ 0.058 & 0.388 $\pm$ 0.050 \\
 & RCV-PCMCI & 21.60 $\pm$ 1.90 & 0.622 $\pm$ 0.034 & 0.416 $\pm$ 0.029 \\
\bottomrule
\end{tabular}
\caption{Performance comparison for Stationary vs Non-stationary datasets}
\label{tab:stationary_nonstationary_performance}
\end{table}

\begin{figure}[htbp]
\centering
\begin{subfigure}[b]{0.48\textwidth}
\centering
\includegraphics[width=\textwidth]{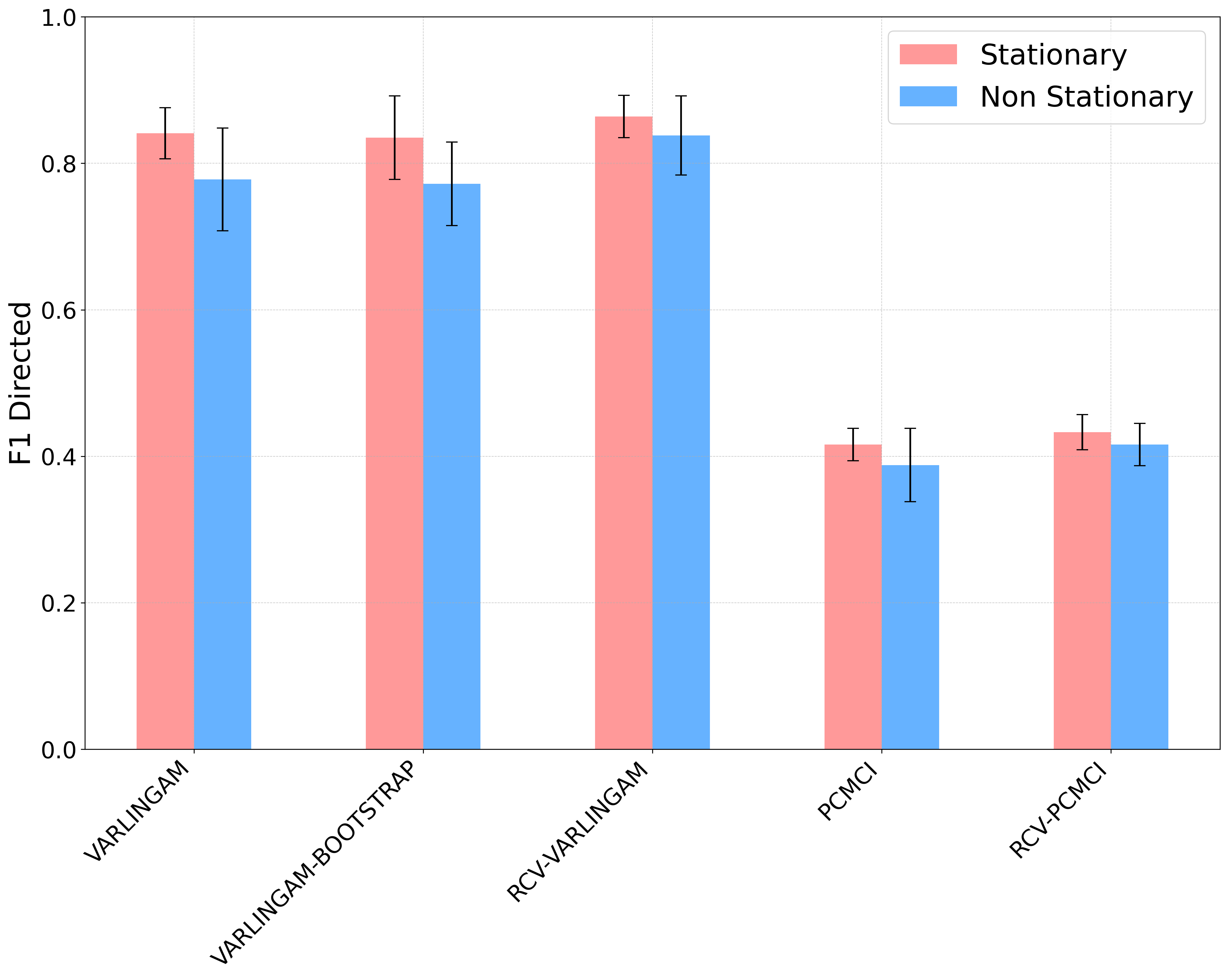}
\caption{F1 Directed Scores}
\label{fig:f1_directed_stationary_nonstationary}
\end{subfigure}
\hfill
\begin{subfigure}[b]{0.48\textwidth}
\centering
\includegraphics[width=\textwidth]{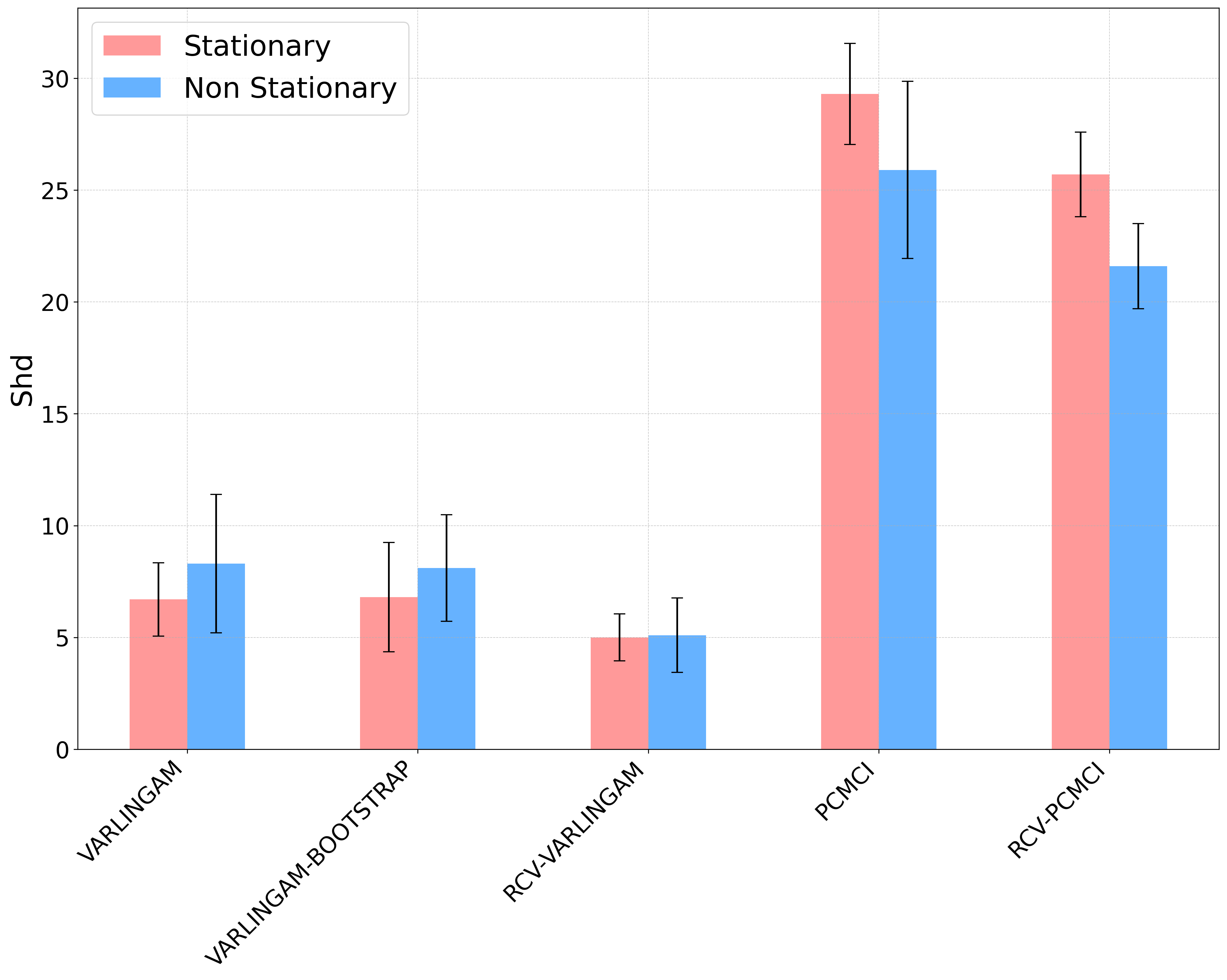}
\caption{Structural Hamming Distance}
\label{fig:shd_stationary_nonstationary}
\end{subfigure}
\caption{Comparison of performance metrics for stationary and non-stationary time series. (a) F1 Directed scores show the accuracy of causal discovery in varying temporal conditions. (b) SHD values indicate the overall structural difference from the true causal graphs in stationary and non-stationary settings.}
\label{fig:stationary_nonstationary_comparison}
\end{figure}

\begin{itemize}
    \item \textbf{Impact of Non-stationarity}: All methods exhibit performance degradation in non-stationary scenarios, confirming the challenges posed by time-varying processes in causal discovery. This aligns with theoretical expectations and underscores the complexity of inferring causal relationships in dynamic systems.

    \item \textbf{RCV-VAR-LiNGAM Resilience}: This method demonstrates superior robustness, maintaining the highest F1 directed scores in both stationary (0.864) and non-stationary (0.838) conditions. The modest decline in performance (3\%) suggests effective mitigation of non-stationarity challenges, possibly due to the cross-validation approach capturing more stable causal relationships.

    \item \textbf{Effectiveness of RCV Approach}: Both RCV variants show improvements over their base methods. RCV-VAR-LiNGAM outperforms VAR-LiNGAM by 2.7\% in stationary and 7.7\% in non-stationary cases. Similarly, RCV-PCMCI improves upon PCMCI, with a notable 7.2\% increase in F1 directed score for non-stationary data.

    \item \textbf{PCMCI Limitations}: PCMCI and its RCV variant struggle considerably in both scenarios, with significantly lower F1 directed scores and higher SHD values. This suggests potential limitations in handling complex temporal dynamics, particularly in non-stationary settings.
\end{itemize}

Figure \ref{fig:stationary_nonstationary_comparison} visually reinforces these findings, clearly illustrating the superior performance of RCV-VAR-LiNGAM across both stationary and non-stationary conditions. The consistent improvement of RCV methods over their base counterparts, particularly in challenging non-stationary scenarios, supports the hypothesis that the RCV approach enhances the robustness of causal discovery methods.

\subsubsection{Sparse vs Dense Causal Structures}

Table \ref{tab:sparse_dense_performance} reveals intriguing patterns in method performance across different causal densities, highlighting the varying effectiveness of causal discovery techniques in sparse and dense networks.

\begin{table}[htbp]
\centering
\small
\begin{tabular}{llccc}
\toprule
Characteristic & Method & SHD & F1 Score & F1\_directed \\
\midrule
\multirow{5}{*}{Sparse} 
 & VAR-LiNGAM & 6.80 $\pm$ 1.23 & 0.768 $\pm$ 0.055 & 0.754 $\pm$ 0.038 \\
 & VL-Bootstrap & 5.80 $\pm$ 1.99 & 0.783 $\pm$ 0.084 & 0.775 $\pm$ 0.076 \\
 & RCV-VAR-LiNGAM & \textbf{3.70 $\pm$ 0.95} & \textbf{0.837 $\pm$ 0.068} & \textbf{0.828 $\pm$ 0.056} \\
 & PCMCI & 13.10 $\pm$ 2.85 & 0.622 $\pm$ 0.049 & 0.598 $\pm$ 0.057 \\
 & RCV-PCMCI & 8.20 $\pm$ 1.23 & 0.649 $\pm$ 0.029 & 0.649 $\pm$ 0.029 \\
\midrule
\multirow{5}{*}{Dense} 
 & VAR-LiNGAM & \textbf{7.90 $\pm$ 3.87} & \textbf{0.870 $\pm$ 0.051} & \textbf{0.859 $\pm$ 0.072} \\
 & VL-Bootstrap & 8.90 $\pm$ 4.38 & 0.865 $\pm$ 0.051 & 0.840 $\pm$ 0.080 \\
 & RCV-VAR-LiNGAM & 10.50 $\pm$ 3.34 & 0.766 $\pm$ 0.098 & 0.766 $\pm$ 0.098 \\
 & PCMCI & 32.50 $\pm$ 1.96 & 0.752 $\pm$ 0.022 & 0.505 $\pm$ 0.029 \\
 & RCV-PCMCI & 24.80 $\pm$ 1.81 & 0.591 $\pm$ 0.037 & 0.443 $\pm$ 0.044 \\
\bottomrule
\end{tabular}
\caption{Performance comparison for Sparse vs Dense datasets}
\label{tab:sparse_dense_performance}
\end{table}

\begin{figure}[htbp]
\centering
\begin{subfigure}[b]{0.48\textwidth}
\centering
\includegraphics[width=\textwidth]{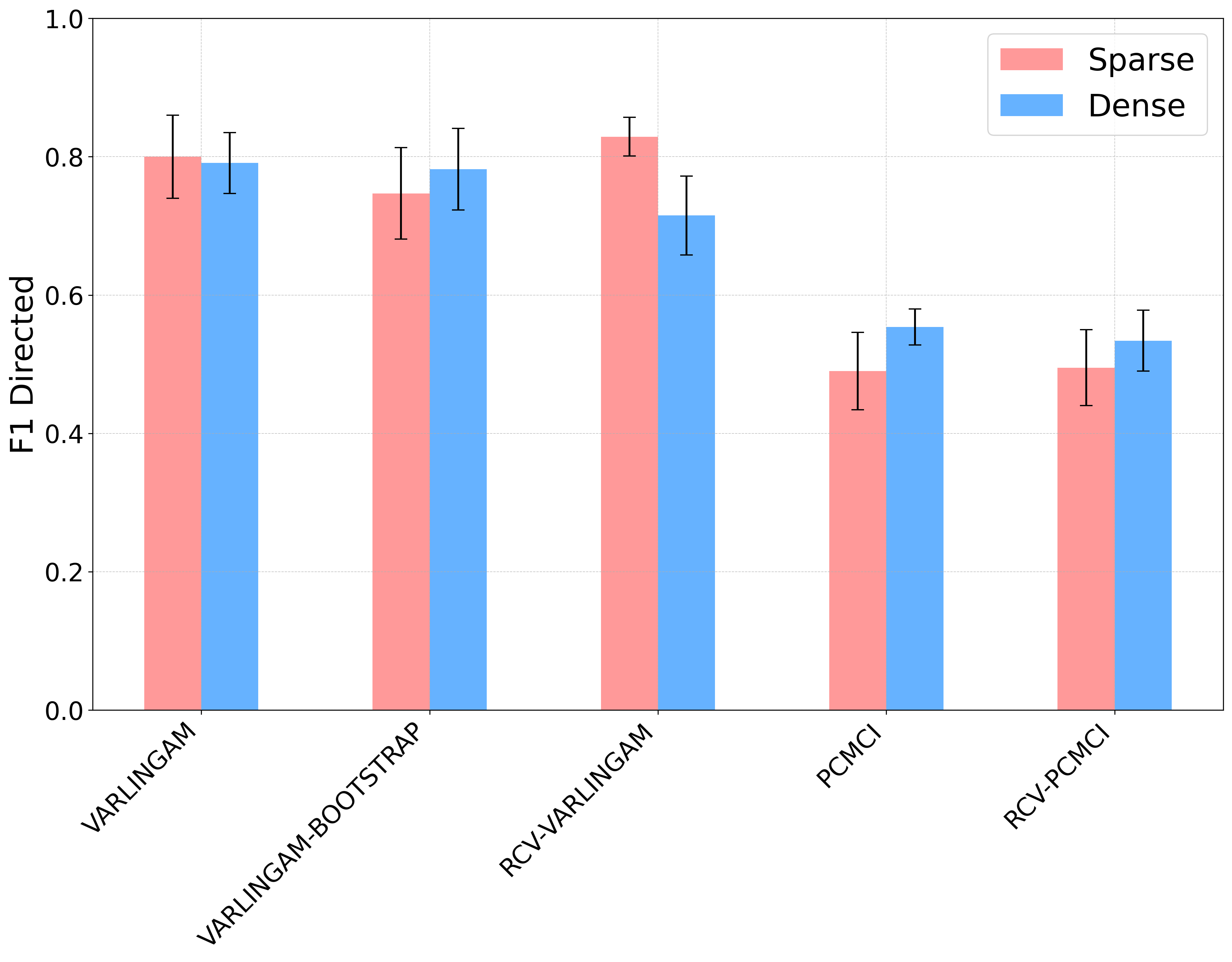}
\caption{F1 Directed Scores}
\label{fig:f1_directed_sparse_dense}
\end{subfigure}
\hfill
\begin{subfigure}[b]{0.48\textwidth}
\centering
\includegraphics[width=\textwidth]{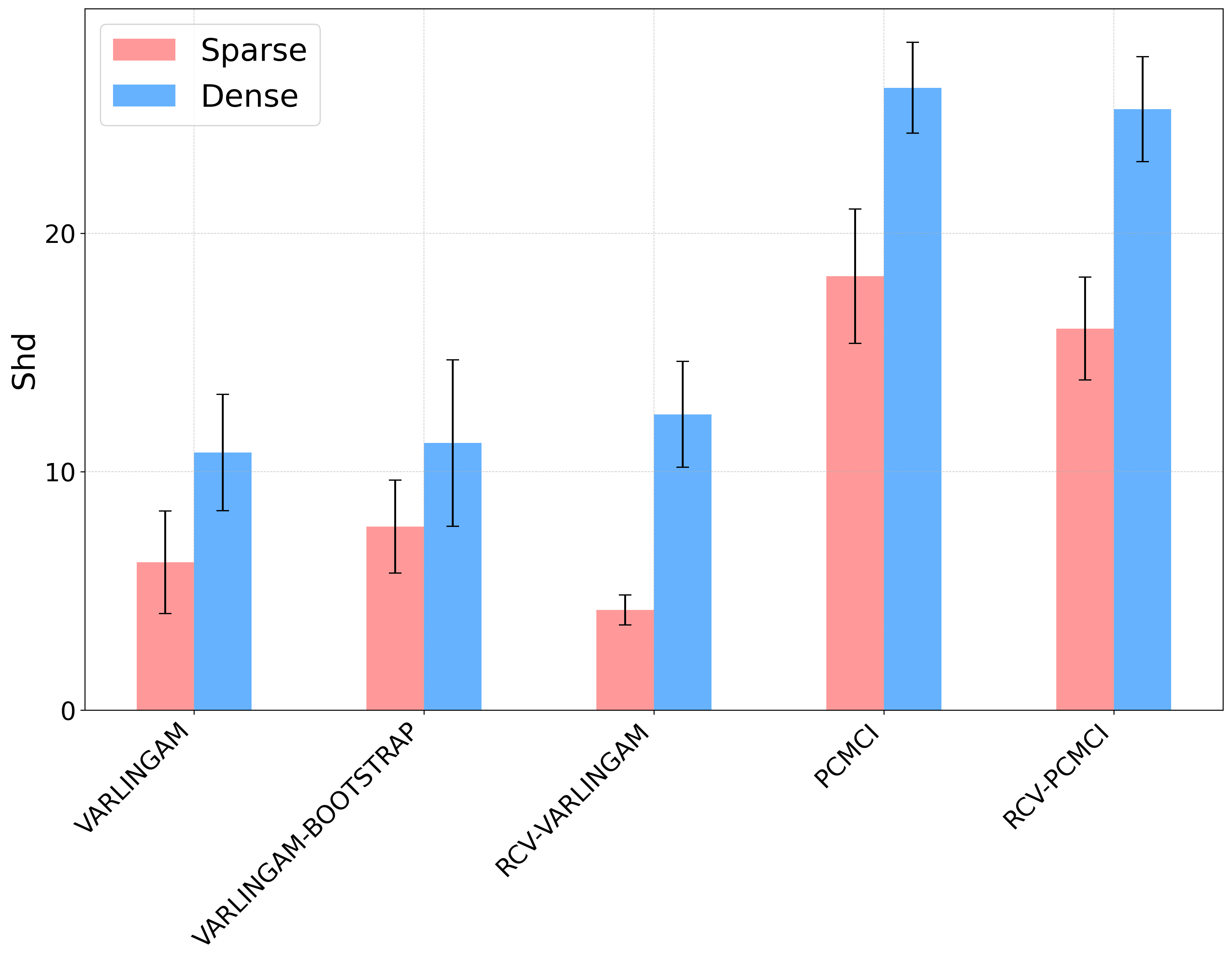}
\caption{Structural Hamming Distance}
\label{fig:shd_sparse_dense}
\end{subfigure}
\caption{Comparison of performance metrics for sparse and dense causal structures. (a) F1 Directed scores show the accuracy of causal discovery in different structural densities. (b) SHD values indicate the overall structural difference from the true causal graphs for sparse and dense networks.} 
\label{fig:sparse_dense_comparison}
\end{figure}

\begin{itemize}
    \item \textbf{Sparse Structure Performance}: RCV-VAR-LiNGAM demonstrates superior performance in sparse causal structures, achieving the highest F1 directed score (0.828) and lowest SHD (3.70). This suggests that the RCV approach is particularly effective in identifying and validating causal relationships in sparsely connected systems.

    \item \textbf{Dense Structure Dynamics}: Interestingly, the performance dynamics shift in dense causal structures. VAR-LiNGAM and its bootstrap variant outperform other methods, with F1 directed scores of 0.859 and 0.840 respectively. This unexpected result suggests that the original VAR-LiNGAM may be more adept at capturing complex, interconnected causal networks.

    \item \textbf{RCV Method Limitations}: While RCV-VAR-LiNGAM excels in sparse scenarios, its performance declines in dense structures (F1 directed: 0.766). This could indicate that the cross-validation approach, while effective for sparse networks, may oversimplify or struggle to capture the complexity of densely connected causal relationships.

    \item \textbf{PCMCI Challenges}: Both PCMCI and RCV-PCMCI demonstrate significant difficulties with dense structures, as evidenced by their high SHD values and low F1 directed scores. This suggests a fundamental limitation in these methods when dealing with highly interconnected causal networks.
\end{itemize}

Figure \ref{fig:sparse_dense_comparison} visually reinforces these observations, clearly illustrating the varying effectiveness of different methods depending on the causal structure's density. The contrasting performance of methods between sparse and dense scenarios underscores the importance of considering network density when selecting appropriate causal discovery techniques.

\subsection{Performance Across Different Dataset Scales}

\subsubsection{Varying Number of Variables}

\begin{figure}[htbp]
\centering
\begin{subfigure}[b]{0.8\textwidth}
\centering\
\includegraphics[width=\textwidth]{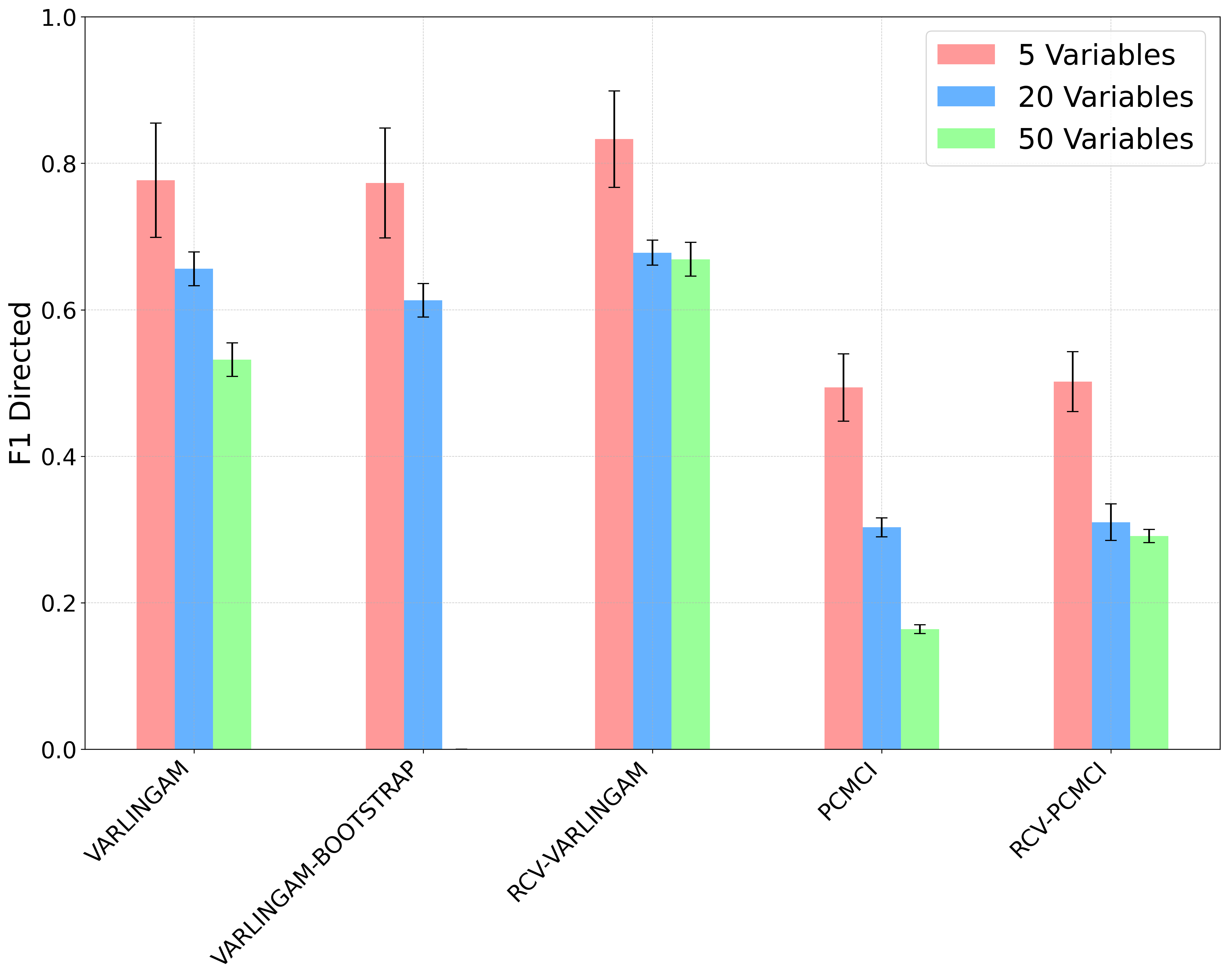}
\caption{F1 Directed Scores}
\label{fig:f1_directed_scale_n_var}
\end{subfigure}

\vspace{0.5cm}

\begin{subfigure}[b]{0.8\textwidth}
\centering
\includegraphics[width=\textwidth]{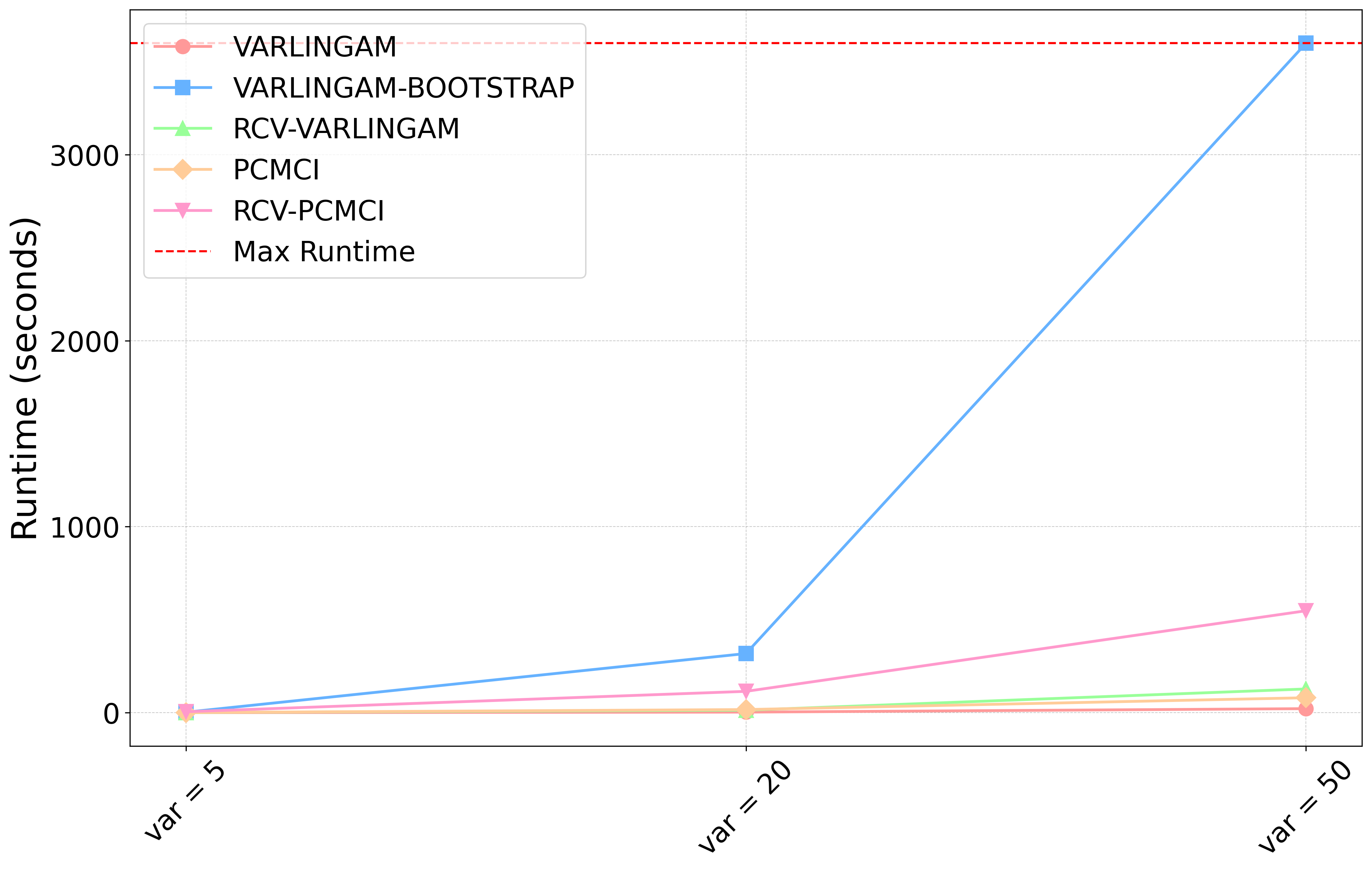}
\caption{Runtime Comparison}
\label{fig:runtime_scale_n_var}
\end{subfigure}
\caption{Comparison of F1 directed scores and runtime for varying number of variables. (a) F1 Directed scores show the accuracy of causal discovery across different variable scales. (b) Runtime comparison illustrates the computational efficiency of each method as the number of variables increases. Note that VL-Bootstrap exceeds the experimental time limit (1 hour or 3600s) for 50 variables, resulting in zero F1 performance and maximum runtime in the plots.}
\label{fig:scale_n_var_comparison}
\end{figure}

Table \ref{tab:scale_n_var_performance} and Figure \ref{fig:scale_n_var_comparison} illustrate the performance and computational efficiency of causal discovery methods as the number of variables increases from 5 to 50.

\begin{table}[htbp]
\centering
\small
\begin{tabular}{llccc}
\toprule
Variables & Method & SHD & F1 Score & F1\_directed \\
\midrule
\multirow{5}{*}{n = 5} 
 & VAR-LiNGAM & 8.80 $\pm$ 2.20 & 0.731 $\pm$ 0.063 & 0.718 $\pm$ 0.060 \\
 & VL-Bootstrap & 7.60 $\pm$ 3.53 & 0.756 $\pm$ 0.084 & 0.756 $\pm$ 0.084 \\
 & RCV-VAR-LiNGAM & \textbf{5.30 $\pm$ 1.06} & \textbf{0.792 $\pm$ 0.039} & \textbf{0.792 $\pm$ 0.039} \\
 & PCMCI & 21.90 $\pm$ 2.51 & 0.560 $\pm$ 0.066 & 0.401 $\pm$ 0.033 \\
 & RCV-PCMCI & 13.60 $\pm$ 1.35 & 0.619 $\pm$ 0.038 & 0.481 $\pm$ 0.044 \\
\midrule
\multirow{5}{*}{n = 20} 
 & VAR-LiNGAM & 162.00 $\pm$ 14.33 & 0.665 $\pm$ 0.018 & 0.656 $\pm$ 0.023 \\
 & VL-Bootstrap & 175.50 $\pm$ 10.88 & 0.632 $\pm$ 0.028 & 0.613 $\pm$ 0.023 \\
 & RCV-VAR-LiNGAM & \textbf{105.80 $\pm$ 3.99} & \textbf{0.680 $\pm$ 0.017} & \textbf{0.678 $\pm$ 0.017} \\
 & PCMCI & 464.20 $\pm$ 18.50 & 0.428 $\pm$ 0.018 & 0.303 $\pm$ 0.013 \\
 & RCV-PCMCI & 215.30 $\pm$ 4.92 & 0.340 $\pm$ 0.027 & 0.310 $\pm$ 0.025 \\
\midrule
\multirow{5}{*}{n = 50} 
 & VAR-LiNGAM & 560.50 $\pm$ 43.21 & 0.540 $\pm$ 0.023 & 0.532 $\pm$ 0.023 \\
 & VL-Bootstrap & \multicolumn{3}{c}{N/A} \\
 & RCV-VAR-LiNGAM & \textbf{237.90 $\pm$ 12.95} & \textbf{0.671 $\pm$ 0.024} & \textbf{0.669 $\pm$ 0.023} \\
 & PCMCI & 1740.30 $\pm$ 67.48 & 0.225 $\pm$ 0.006 & 0.164 $\pm$ 0.006 \\
 & RCV-PCMCI & 478.90 $\pm$ 30.00 & 0.301 $\pm$ 0.012 & 0.291 $\pm$ 0.009 \\
\bottomrule
\end{tabular}
\caption{Performance comparison for varying number of variables}
\label{tab:scale_n_var_performance}
\end{table}

\begin{itemize}
    \item \textbf{RCV-VAR-LiNGAM Superiority}: This method consistently outperforms others across all scales, maintaining the highest F1 directed scores (0.792, 0.678, and 0.669 for n = 5, 20, and 50 respectively). Its superior performance is particularly evident in high-dimensional settings, demonstrating robust scalability.

    \item \textbf{Performance Degradation}: All methods exhibit a decline in performance as the number of variables increases, with PCMCI showing the most significant drop (F1 directed from 0.401 to 0.164 as n increases from 5 to 50). This highlights the challenges of causal discovery in high-dimensional spaces.

    \item \textbf{Computational Efficiency Trade-offs}: While VAR-LiNGAM shows the fastest runtime, RCV-VAR-LiNGAM offers a balanced trade-off between performance and computational cost. Notably, the bootstrap method becomes computationally infeasible for n = 50, emphasising the importance of efficient algorithms for large-scale analyses.

    \item \textbf{PCMCI Limitations}: Both PCMCI and RCV-PCMCI struggle with increasing dimensionality, suggesting these methods may be less suitable for high-dimensional causal discovery tasks.
\end{itemize}

Figure \ref{fig:scale_n_var_comparison} visually reinforces these findings, clearly illustrating RCV-VAR-LiNGAM's consistent performance across different scales and the varying computational demands of each method.

\subsubsection{Varying Time Series Length}
\begin{figure}[htbp]
\centering
\begin{subfigure}[b]{0.8\textwidth}
\centering\
\includegraphics[width=\textwidth]{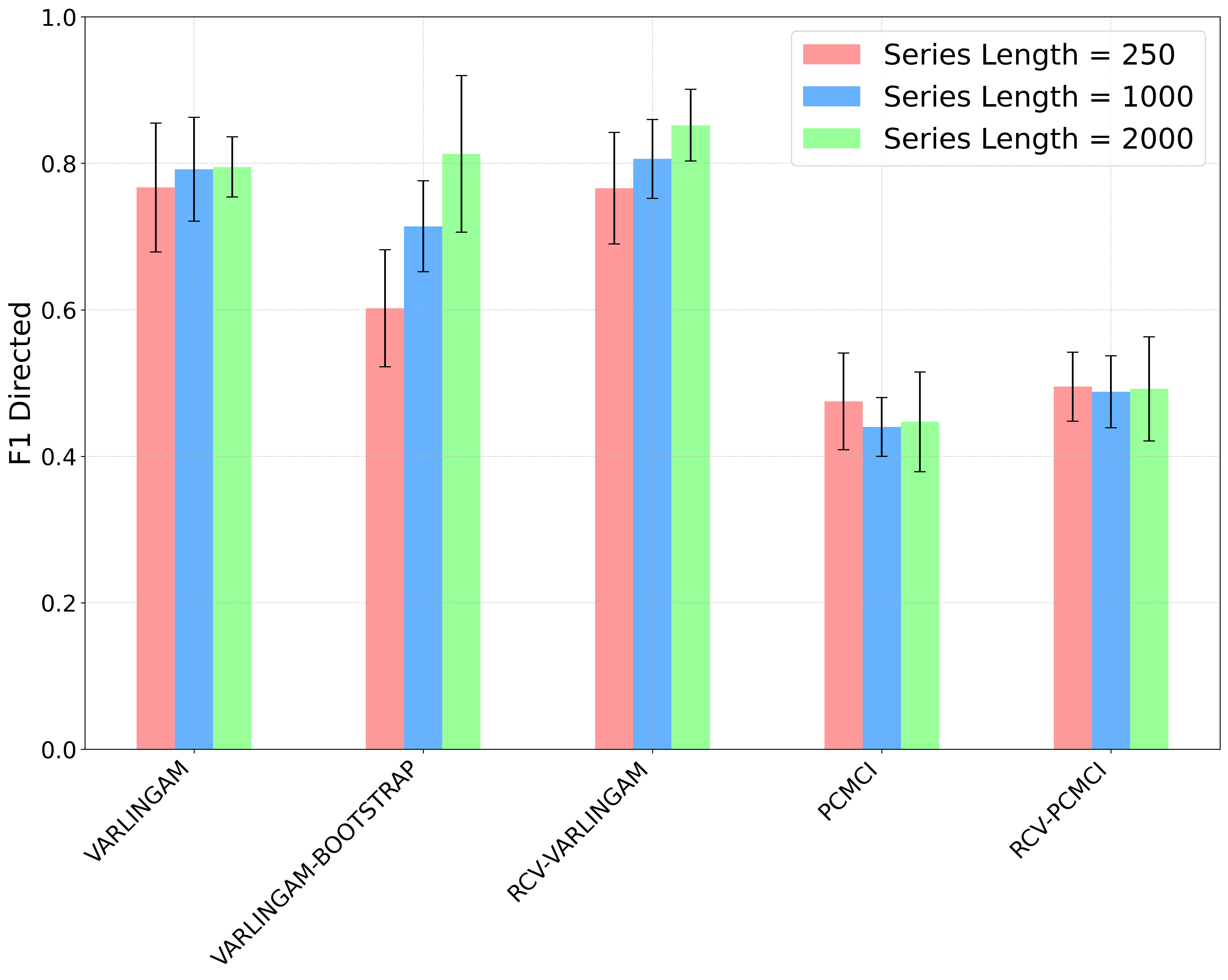}
\caption{F1 Directed Scores}
\label{fig:f1_directed_scale_t}
\end{subfigure}

\vspace{0.5cm}

\begin{subfigure}[b]{0.8\textwidth}
\centering
\includegraphics[width=\textwidth]{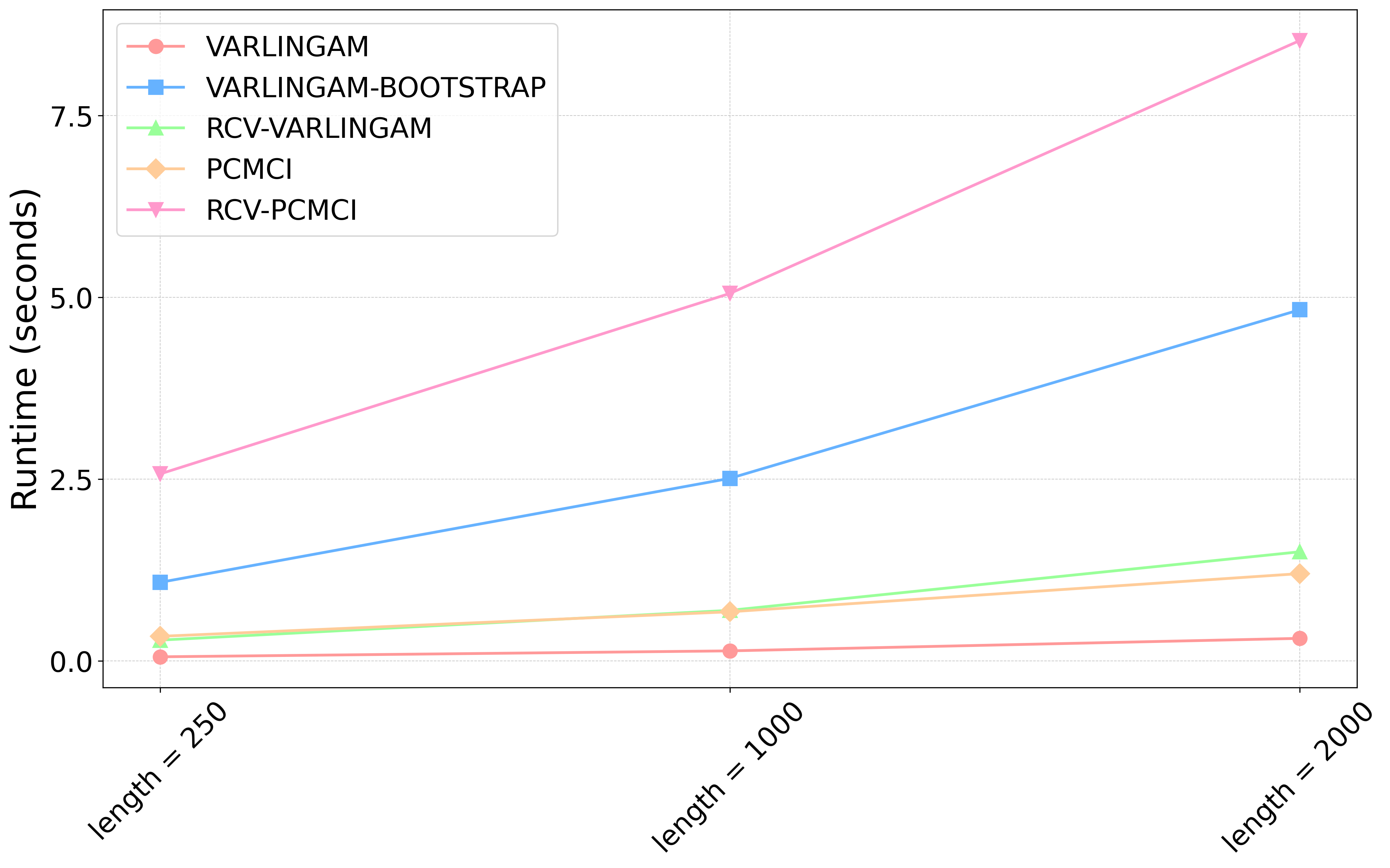}
\caption{Runtime Comparison}
\label{fig:runtime_scale_t}
\end{subfigure}
\caption{Comparison of F1 directed scores and runtime for varying time series lengths. (a) F1 Directed scores demonstrate the accuracy in identifying causal relations as the time series length increases. (b) Runtime comparison illustrates the computational efficiency of each method for different time series lengths. Note that all methods complete within the experimental time limit, showing varying degrees of runtime increase with longer time series.}
\label{fig:scale_t_comparison}
\end{figure}

Table \ref{tab:scale_t_performance} and Figure \ref{fig:scale_t_comparison} illustrate the performance and computational efficiency of causal discovery methods as the time series length increases from 250 to 2000 time points.

\begin{table}[htbp]
\centering
\small
\begin{tabular}{llccc}
\toprule
Length & Method & SHD & F1 Score & F1\_directed \\
\midrule
\multirow{5}{*}{T = 250} 
 & VAR-LiNGAM & 5.90 $\pm$ 1.91 & \textbf{0.771 $\pm$ 0.063} & \textbf{0.771 $\pm$ 0.063} \\
 & VL-Bootstrap & 8.10 $\pm$ 1.79 & 0.722 $\pm$ 0.060 & 0.714 $\pm$ 0.058 \\
 & RCV-VAR-LiNGAM & \textbf{5.20 $\pm$ 1.03} & 0.765 $\pm$ 0.051 & 0.765 $\pm$ 0.051 \\
 & PCMCI & 14.90 $\pm$ 2.02 & 0.477 $\pm$ 0.045 & 0.456 $\pm$ 0.031 \\
 & RCV-PCMCI & 10.00 $\pm$ 1.15 & 0.526 $\pm$ 0.069 & 0.518 $\pm$ 0.063 \\
\midrule
\multirow{5}{*}{T = 1000} 
 & VAR-LiNGAM & 7.70 $\pm$ 2.50 & 0.762 $\pm$ 0.063 & 0.762 $\pm$ 0.063 \\
 & VL-Bootstrap & 6.20 $\pm$ 2.39 & 0.791 $\pm$ 0.070 & 0.791 $\pm$ 0.070 \\
 & RCV-VAR-LiNGAM & \textbf{4.20 $\pm$ 1.99} & \textbf{0.831 $\pm$ 0.088} & \textbf{0.831 $\pm$ 0.088} \\
 & PCMCI & 19.00 $\pm$ 3.33 & 0.509 $\pm$ 0.056 & 0.427 $\pm$ 0.041 \\
 & RCV-PCMCI & 12.30 $\pm$ 1.42 & 0.499 $\pm$ 0.044 & 0.482 $\pm$ 0.026 \\
\midrule
\multirow{5}{*}{T = 2000} 
 & VAR-LiNGAM & 7.10 $\pm$ 1.60 & 0.772 $\pm$ 0.050 & 0.772 $\pm$ 0.050 \\
 & VL-Bootstrap & 6.10 $\pm$ 3.14 & 0.796 $\pm$ 0.093 & 0.796 $\pm$ 0.093 \\
 & RCV-VAR-LiNGAM & \textbf{4.50 $\pm$ 1.84} & \textbf{0.833 $\pm$ 0.066} & \textbf{0.833 $\pm$ 0.066} \\
 & PCMCI & 21.20 $\pm$ 2.86 & 0.527 $\pm$ 0.069 & 0.412 $\pm$ 0.056 \\
 & RCV-PCMCI & 13.30 $\pm$ 0.95 & 0.493 $\pm$ 0.037 & 0.478 $\pm$ 0.035 \\
\bottomrule
\end{tabular}
\caption{Performance comparison for varying time series length}
\label{tab:scale_t_performance}
\end{table}

\begin{itemize}
    \item \textbf{RCV-VAR-LiNGAM Consistency}: This method demonstrates superior performance when applying for longer time series lengths, achieving the highest F1 directed scores in the two longer datasets (0.831 and 0.833 for T = 1000 and 2000 respectively). Its ability to effectively leverage additional temporal information is evident in the steady improvement of performance with increased time series length.

    \item \textbf{Performance Improvement Trend}: Generally, all methods show improved performance as the time series length increases, suggesting that longer time series provide more information for accurate causal discovery. However, the extent of improvement varies among methods.

    \item \textbf{PCMCI Limitations}: Both PCMCI and RCV-PCMCI show limited improvement with increasing time series length, with F1 directed scores remaining relatively flat (0.456 to 0.412 for PCMCI, 0.518 to 0.478 for RCV-PCMCI). This suggests these methods may be less effective at utilising additional temporal data.

    \item \textbf{Computational Efficiency Trade-offs}: While VAR-LiNGAM maintains the fastest runtime, RCV-VAR-LiNGAM offers a balanced trade-off between performance and computational cost. The bootstrap method (VL-Bootstrap) shows the steepest increase in runtime, potentially limiting its applicability for very long time series.
\end{itemize}

Figure \ref{fig:scale_t_comparison} visually reinforces these findings, clearly illustrating RCV-VAR-LiNGAM's consistent performance improvement and the varying computational demands of each method as time series length increases.

\section{RCV Method Advantages}

Our proposed RCV methods (RCV-VAR-LiNGAM and RCV-PCMCI) demonstrate several advantages:

\begin{itemize}
    \item \textbf{Robustness to Non-linearity:} RCV-VAR-LiNGAM maintained high performance in non-linear scenarios (F1\_directed: 0.819 $\pm$ 0.052) compared to its base method (0.712 $\pm$ 0.065).
    
    \item \textbf{Noise Resilience:} Both RCV methods showed improved performance in non-Gaussian noise settings. RCV-VAR-LiNGAM achieved an F1\_directed score of 0.818 $\pm$ 0.039 compared to VAR-LiNGAM's 0.758 $\pm$ 0.061.
    
    \item \textbf{Non-stationary Performance:} RCV methods demonstrated better handling of non-stationary data. RCV-VAR-LiNGAM's F1\_directed score (0.838 $\pm$ 0.054) surpassed VAR-LiNGAM (0.778 $\pm$ 0.070) in non-stationary scenarios.
    
    \item \textbf{Scalability:} RCV methods completed all analyses, including larger datasets, while bootstrap methods faced computational limitations.
    
    \item \textbf{Consistency:} RCV-VAR-LiNGAM showed the most consistent performance across various data characteristics and scales.
\end{itemize}

These improvements suggest that our RCV methods offer more reliable and versatile causal discovery in diverse time series scenarios.

\section{Application to Simulated fMRI Data}
To validate our methods on complex, neuroimaging-like data, we applied all five causal discovery techniques to the simulated fMRI dataset. While not real-world data, this dataset provides a challenging benchmark with known ground truth causal structures, allowing for accurate performance evaluation. Table \ref{tab:fmri_results} presents the results of this analysis.

\begin{table}[htbp]
\centering
\small
\begin{tabular}{lcccc}
\toprule
Method & SHD & F1 Score & F1\_directed & Runtime (s) \\
\midrule
VAR-LiNGAM & 16.444 $\pm$ 9.040 & 0.484 $\pm$ 0.067 & 0.462 $\pm$ 0.071 & \textbf{0.174} \\
VL-Bootstrap & 21.704 $\pm$ 16.877 & 0.438 $\pm$ 0.069 & 0.408 $\pm$ 0.072 & 4.765 \\
RCV-VAR-LiNGAM & \textbf{11.333 $\pm$ 5.884} & \textbf{0.508 $\pm$ 0.140} & \textbf{0.506 $\pm$ 0.140} & 0.7835 \\
PCMCI & 31.037 $\pm$ 22.346 & 0.373 $\pm$ 0.089 & 0.367 $\pm$ 0.081 & 0.526 \\
RCV-PCMCI & 25.444 $\pm$ 18.260 & 0.393 $\pm$ 0.120 & 0.393 $\pm$ 0.120 & 3.837 \\
\bottomrule
\end{tabular}
\caption{Performance comparison of causal discovery methods on fMRI data}
\label{tab:fmri_results}
\end{table}

The results from the simulated fMRI data analysis reveal several important insights:

\begin{itemize}
    \item \textbf{RCV-VAR-LiNGAM Superiority}: This method demonstrates the best overall performance, achieving the highest F1 and F1\_directed scores (0.508 and 0.506 respectively) and the lowest SHD (11.333). This suggests RCV-VAR-LiNGAM's effectiveness in identifying causal relationships in complex, non-linear time series data similar to those found in neuroimaging.
    
    \item \textbf{Consistency with Synthetic Data}: The relative performance of methods on fMRI data aligns with our findings from synthetic data. RCV methods, particularly RCV-VAR-LiNGAM, outperform their base counterparts, reinforcing their robustness in scenarios mimicking real-world complexity.
    
    \item \textbf{Computational Efficiency Trade-offs}: VAR-LiNGAM shows the fastest runtime (0.174s), while RCV-VAR-LiNGAM offers a good balance between performance and efficiency (0.784s). The bootstrap method (VL-Bootstrap) is significantly slower, potentially limiting its applicability in large-scale fMRI studies.
    
    \item \textbf{High Variability in Results}: The large standard deviations across all methods, particularly in SHD, indicate significant variability in identified causal structures. This highlights the challenging nature of causal discovery in complex neuroimaging-like data.
    
    \item \textbf{PCMCI Limitations}: Both PCMCI and RCV-PCMCI show lower performance on fMRI data compared to other methods, suggesting they may be less suited for this type of complex, non-linear time series analysis.
\end{itemize}

These findings underscore RCV-VAR-LiNGAM's potential for causal discovery in complex time series data, while also highlighting the challenges inherent in analysing intricate, non-linear causal structures similar to those found in brain connectivity patterns. The results emphasise the importance of method selection in causal inference tasks, particularly when dealing with complex, high-dimensional data typical in fields like neuroscience.

\section{Discussion}
Our comprehensive evaluation, encompassing synthetic datasets with varying characteristics and a complex simulated fMRI dataset, reveals several key insights into the performance and applicability of causal discovery methods:

\begin{itemize}
    \item \textbf{RCV-VAR-LiNGAM Superiority:} Consistently outperforming other methods across diverse scenarios, RCV-VAR-LiNGAM demonstrates:

    \begin{itemize}
        \item Enhanced robustness to non-linearity, maintaining high performance in both linear and non-linear settings.
        \item Superior handling of different noise distributions, particularly excelling in non-Gaussian scenarios.
        \item Resilience to non-stationarity, a common challenge in real-world time series data.
        \item Consistent performance across varying causal densities, though with a slight advantage in sparse structures.
    \end{itemize}
    
    \item \textbf{Scalability and Efficiency:} RCV-VAR-LiNGAM exhibits strong scalability, maintaining high performance as the number of variables increases. It offers a balanced trade-off between accuracy and computational cost, particularly evident in its handling of the complex fMRI dataset.

    \item \textbf{Temporal Data Utilisation:} This approach efficiently utilizes additional temporal information for better results as the length of the time series being analyzed increases. This is indicative of its applicability to long-term time series studies, for instance, in domains such as stock market forecasts and climate science.

    \item \textbf{PCMCI Limitations:} Both PCMCI and RCV-PCMCI demonstrate significant challenges in dealing with complicated, dynamic relationships with a multitude of variables. Their performance degradation in these scenarios suggests they could be more suitable for systems with less complexity.

    \item \textbf{Robustness to Data Characteristics:} RCV-VAR-LiNGAM's consistent performance over a range of data characteristics, such as linearity, noise distribution, stationarity, and so on, indicates its wide applicability in causal discovery under diverse scenarios of complex settings, which are typical of real-world investigations.

    \item \textbf{fMRI Data Performance:} While RCV-VAR-LiNGAM shows the best performance on simulated fMRI data, the overall low F1 scores across all methods highlight the challenging nature of causal discovery in neuroimaging-like data. This suggests a need for further refinement in handling such complex, non-linear time series.
\end{itemize}

In conclusion, our proposed RCV methods, particularly RCV-VAR-LiNGAM, offer significant advancements in causal discovery for complex time series data. Their robust performance across various scenarios, from synthetic datasets to simulated fMRI data, positions them as valuable tools for researchers analysing intricate time series in fields such as neuroscience, finance, and climate science. As we continue to make improvements in these methods, their potential impact on causal inference and analysis of complex systems' dynamics and decision-making activities in real-world applications is significant and could lead to a more accurate and trustworthy inference.

\chapter{Enhanced ABM Validation Framework}

This chapter introduces the third major contribution of our work: a Context-Aware ABM Validation Framework. We've built upon and expanded the work of Guerini et al. \cite{Guerini2017} to create an advanced framework that tackles key shortcomings in existing ABM validation methods. This framework allows for a flexible selection of causal inference methods, introduces new performance metrics, and provides a comprehensive approach to data organization, analysis, and causal structure detection. Our goal is to make ABM validation more accurate and reliable, especially for complex systems like those found in finance.

\section{Overview}
We've developed this improved framework for validating Agent-Based Models (ABMs) by enhancing the work of Guerini and Moneta (2017) \cite{Guerini2017}. Our approach integrates advanced causal discovery techniques to address the complexities of validating ABMs against real-world data. The framework comprises four key stages: (1) dataset uniformity, (2) property analysis, (3) causal structure identification, and (4) validation assessment.
\\ \\
By combining the strengths of the original framework with our enhanced causal discovery techniques, including our new Robust Cross-Validated (RCV) methods, we aim to make ABM validation in complex systems more reliable and easier to interpret. This improvement helps us assess more accurately how well ABMs reflect real-world phenomena. As a result, researchers in finance and economics would have a more robust tool for validating their models.

\section{Causal Discovery}
A key extension in our framework is the integration of multiple causal discovery methods, including traditional approaches and our novel RCV extensions. The framework supports:

\begin{itemize}
\item VAR-LiNGAM (Vector Autoregressive Linear Non-Gaussian Acyclic Model)
\item PCMCI (Peter and Clark Momentary Conditional Independence)
\item RCV-VAR-LiNGAM (our proposed method)
\item RCV-PCMCI (our proposed method)
\item VAR-LiNGAM with Bootstrap
\end{itemize}

This wide range of methods gives researchers the flexibility to choose the most suitable technique for their needs, such as data characteristics, research goals, and computational limitations they might have. For projects where efficiency is crucial, traditional methods like VAR-LiNGAM offer faster processing. On the other hand, our RCV methods provide enhanced robustness, though they require more computational power. By including our RCV methods, we're addressing key weaknesses in existing approaches, particularly in handling noise and non-stationary behaviours which are common in financial time series, offering a trade-off between computational efficiency and result reliability. This means researchers can choose between faster processing and more reliable results, depending on what's more important for their specific project.

\section{Components}

\subsection{Data Preprocessing and Uniformity}
The initial stage of our framework focuses on ensuring data consistency and compatibility between real-world and ABM-generated datasets. This process involves handling missing values and outliers, applying necessary transformations (e.g., logarithmic, percentage), and aligning time series lengths.

\subsection{Property Analysis}
Following preprocessing, our framework conducts a comprehensive analysis of key statistical properties for both real-world and ABM data. This analysis includes:

\begin{itemize}
\item Linearity assessment
\item Stationarity testing
\item Evaluation of statistical equilibrium (for ABM data)
\item Ergodicity analysis (for ABM data)
\end{itemize}

These properties provide critical insights into the fundamental characteristics of the data, informing the selection of appropriate causal discovery methods and aiding in the interpretation of validation results.

\subsection{Causal Identification}
The core of our framework lies in its flexible approach to causal structure identification. By incorporating multiple methods, including our novel RCV approaches, we enable a more nuanced and robust analysis of causal relationships. This approach allows for easy switching between different causal discovery techniques, facilitating comparative analyses and method-specific optimizations.

\begin{figure}[htbp]
    \centering
    \includegraphics[width=1.0\textwidth]{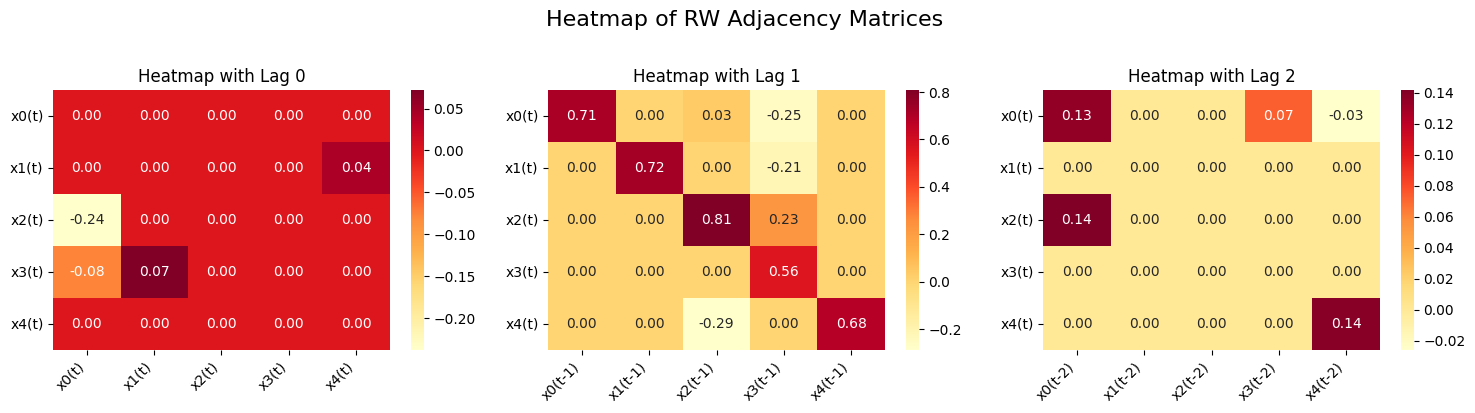}
    \caption{Heatmap of adjacency matrix identified by the framework}
    \label{fig:heatmap}
\end{figure}

\begin{figure}[htbp]
    \centering
    \includegraphics[width=0.8\textwidth]{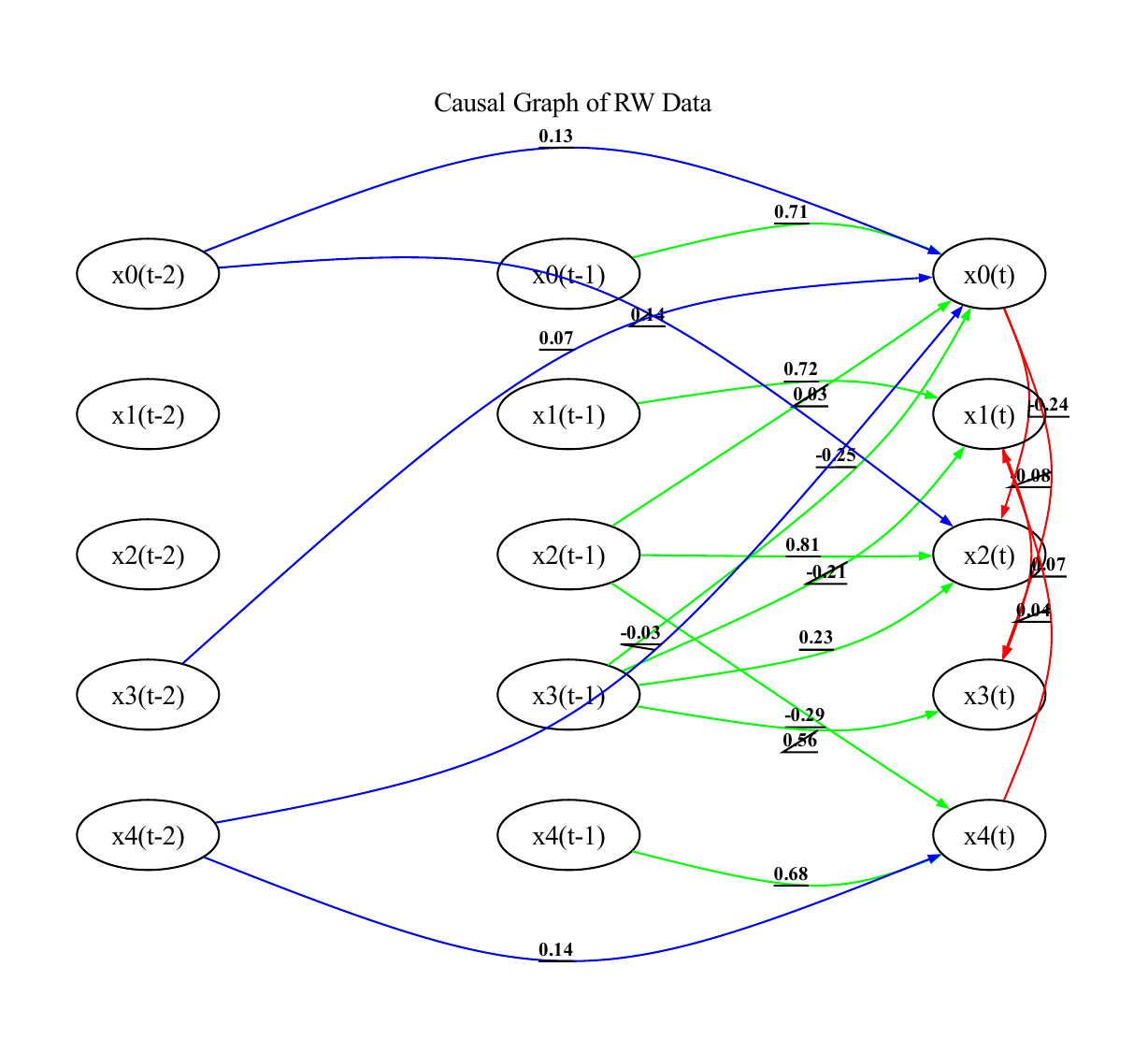}
    \caption{Causal graph representation of the structure identified by the framework.}
    \label{fig:causalgraph}
\end{figure}

Our framework generates two key visualizations of the identified causal structure. Figure \ref{fig:heatmap} presents a heatmap of the adjacency matrix, where darker colors indicate stronger causal relationships, providing an intuitive overview of causal strengths between variables. Figure \ref{fig:causalgraph} displays the same structure as a directed graph, with edge colours representing the lag number of the cause, clearly showing the causal network's topology. These complementary visualizations, applicable to both real-world and ABM-generated data, allow for direct comparison of causal structures. The heatmap excels at showing the overall distribution of relationships, while the graph is useful for tracing specific causal pathways, collectively aiding in assessing an ABM's ability to replicate observed causal dynamics.

\subsection{Validation Assessment}
The final stage involves a multi-faceted assessment of model validity. We employ a range of metrics to compare the causal structures identified in real-world and ABM data, combining methods used in evaluating causal discovery performance with similarity tests used in Guerini and Moneta (2017). This comprehensive approach provides a nuanced view of model performance, capturing both structural and quantitative aspects of causal relationship alignment.
\\ \\
Our metrics include:
\begin{enumerate}
    \item Structural Hamming Distance (SHD): Measures the structural differences between the identified causal graphs.
    \item F1 Score (undirected and directed): Evaluates the precision and recall of the identified causal relationships.
    \item Frobenius norm: Quantifies the overall difference between adjacency matrices.
    \item Similarity tests: Adapted from Guerini and Moneta (2017), these include:
        \\
        a) Sign similarity:
        \begin{equation}
        S_{sign} = \frac{1}{n^2} \sum_{i,j} I(\text{sign}(A_{ij}) = \text{sign}(B_{ij}))
        \end{equation}
        
        b) Size similarity:
        \begin{equation}
        S_{size} = \frac{1}{n^2} \sum_{i,j} I(|A_{ij} - B_{ij}| \leq 2\sigma_{A_{ij}})
        \end{equation}
        
        c) Conjugate similarity:
        \begin{equation}
        S_{conj} = \frac{1}{n^2} \sum_{i,j} I(\text{sign}(A_{ij}) = \text{sign}(B_{ij})) \cdot I(|A_{ij} - B_{ij}| \leq 2\sigma_{A_{ij}})
        \end{equation}
        
        Where $A$ and $B$ are the adjacency matrices for the real-world and ABM causal structures respectively, $n$ is the number of variables, $I(\cdot)$ is the indicator function, and $\sigma_{A_{ij}}$ is the standard deviation of $A_{ij}$ across multiple ABM simulations.
\end{enumerate}

These similarity measures provide a more detailed comparison of the causal structures, allowing for a fine-grained analysis of how well the ABM captures the causal relationships present in the real-world data. The sign similarity assesses the directional accuracy of the causal links, the size similarity evaluates the magnitude of the causal effects, and the conjugate similarity combines both aspects.
\\ \\
By incorporating these diverse metrics, our framework provides a comprehensive assessment of ABM validity, encompassing both the structural accuracy of the causal relationships and the quantitative similarity of the causal effects. This approach allows researchers to gain deeper insights into the strengths and limitations of their ABMs in replicating real-world causal structures.

\section{Preliminary Evaluation}
We conducted a preliminary experiment to provide an initial assessment of our enhanced ABM validation framework, particularly focusing on the incorporation of our proposed RCV approach. While this experiment is simplified, it offers valuable insights into the framework's potential effectiveness.

\subsection{Experimental Setup}
We generated 10 datasets using the dataset generator described in Chapter 4, all sharing the same underlying causal structure (identical adjacency matrices). This setup mimics a scenario where an ABM perfectly replicates the causal relationships of the real-world system it models. The experimental procedure was as follows:

\begin{itemize}
    \item One dataset was designated as the `real-world data'.

    \item The remaining nine datasets were labelled as `ABM data\_0' through `ABM data\_8'.

    \item Ideally, the estimated causal structures should exhibit 100\% similarity after discovery.
\end{itemize}

We ran the validation framework twice, once using the original VAR-LiNGAM method and once using our proposed RCV-VAR-LiNGAM method during the Causal Identification step.

\subsection{Results and Analysis}
Table \ref{tab:framework_evaluation} presents the results of our preliminary evaluation, comparing the performance of VAR-LiNGAM and RCV-VAR-LiNGAM within our framework.

\begin{table}[htbp]
\centering
\begin{tabular}{lcc}
\hline
Metric & VAR-LiNGAM & RCV-VAR-LiNGAM \\
\hline
SHD & 26.667 $\pm$ 4.796 & \textbf{10.111 $\pm$ 3.060} \\
F1 Score & 0.492 $\pm$ 0.056 & \textbf{0.659 $\pm$ 0.076} \\
F1 directed & 0.488 $\pm$ 0.064 & \textbf{0.659 $\pm$ 0.076} \\
Frobenius Norm & 0.649 $\pm$ 0.144 & \textbf{0.494 $\pm$ 0.094} \\
Sign Similarity & 0.720 $\pm$ 0.026 & \textbf{0.890 $\pm$ 0.041} \\
Size Similarity & 0.998 $\pm$ 0.007 & \textbf{1.000 $\pm$ 0.000} \\
Conjugate Similarity & 0.720 $\pm$ 0.026 & \textbf{0.890 $\pm$ 0.041} \\
Runtime (s) & \textbf{1.103} & 5.160 \\
\hline
\end{tabular}
\caption{Comparison of Framework Performance: VAR-LiNGAM vs RCV-VAR-LiNGAM}
\label{tab:framework_evaluation}
\end{table}

The results suggest that our framework, when utilising the RCV approach, yields estimates closer to the ideal accuracy. Key observations include:

\begin{itemize}
    \item \textbf{Structural Accuracy:} RCV-VAR-LiNGAM achieved a significantly lower Structural Hamming Distance (SHD) (10.111 vs 26.667), indicating a more accurate structural estimation of the causal graph. This substantial improvement suggests that the RCV approach is more effective in capturing the true causal relationships.
    
    \item \textbf{Relationship Identification:} Both undirected and directed F1 scores improved with RCV-VAR-LiNGAM (0.659 vs 0.492), suggesting better precision and recall in identifying causal relationships. This enhancement in F1 scores indicates that the RCV method is more adept at correctly identifying both the presence and direction of causal links.
    
    \item \textbf{Similarity Measures:} RCV-VAR-LiNGAM showed higher sign and conjugate similarities (0.890 vs 0.720 for both). This means it's better at identifying the direction and strength of causal relationships. It also achieved perfect size similarity (1.000), showing its accuracy in estimating the strength of these relationships.
    
    \item \textbf{Overall Difference:} The lower Frobenius Norm (0.494 vs 0.649) indicates that RCV-VAR-LiNGAM's estimated causal structures were closer to the true structures. This further supports its superior performance.
    
    \item \textbf{Computational Trade-off:} RCV-VAR-LiNGAM took about 5 times longer to run (5.160s vs 1.103s). This represents a trade-off between accuracy and speed. The longer runtime is due to the cross-validation process, but the significant accuracy improvements may justify this.
\end{itemize}

These results highlight the potential of our enhanced framework, especially when using the RCV-VAR-LiNGAM method. It consistently outperforms across multiple metrics, suggesting it offers a more robust and precise estimation of causal structures. This is crucial for accurately assessing how well ABMs reflect real-world systems. However, the increased computational cost should be considered when applying this method to larger or more complex datasets.

\section{Summary}
Our enhanced ABM validation framework offers several key improvements:

\begin{enumerate}
    \item \textbf{Improved Accuracy:} The initial evaluation shows that using the RCV approach leads to more accurate identification of causal structures across multiple metrics.

    \item \textbf{Flexibility:} The framework allows researchers to choose from various causal discovery methods based on their specific needs and data characteristics.

    \item \textbf{Comprehensive Validation:} By integrating multiple performance metrics, the framework provides a well-rounded assessment of model validity.

    \item \textbf{Efficiency-Accuracy Trade-off:} While the RCV method improves accuracy, it takes more computational time. This allows users to balance precision with resource constraints.
\end{enumerate}

This framework addresses challenges in assessing how well ABMs reflect real-world phenomena, particularly in complex systems like financial markets. It offers researchers the tools to adapt the validation process to their specific goals and data types, potentially leading to more reliable and insightful ABMs.
\\ \\
However, it's important to note the limitations of our initial evaluation. The experiment uses a simplified scenario and a limited number of datasets. More thorough testing, including different causal structures, data characteristics, and larger sample sizes, would be needed to fully validate the framework's effectiveness across diverse scenarios.
\\ \\
As we apply this framework to various fields, we expect it will contribute to the ongoing improvements of ABMs, ultimately supporting more informed decision-making in complex systems analysis.

\chapter{Conclusion}

\section{Summary of Key Findings and Addressing Research Challenges}
This study has made significant strides in enhancing causal discovery methods for time series data, with particular applications in finance and other complex domains. Our key findings and contributions directly address the three main research challenges identified at the outset:

\begin{enumerate}
    \item \textbf{Enhancing Robustness of Causal Discovery Methods:} We developed and implemented Robust Cross-Validated (RCV) versions of VAR-LiNGAM and PCMCI. These methods demonstrated improved performance across various scenarios:
    
    \begin{itemize}
        \item RCV-VAR-LiNGAM consistently outperformed existing methods in both linear and non-linear scenarios, with F1 directed scores of 0.848 and 0.819 respectively.
        
        \item It showed superior noise resilience, particularly in non-Gaussian settings (F1 directed score of 0.818 compared to VAR-LiNGAM's 0.758).
        
        \item The method demonstrated better handling of non-stationary data, with an F1 directed score of 0.838 in non-stationary scenarios compared to VAR-LiNGAM's 0.778.
    \end{itemize}
    
    \item \textbf{Understanding Dataset Characteristics' Impact:} Through extensive testing on synthetic datasets, we provided insights into how various factors affect the performance of causal discovery methods:
    \begin{itemize}
        \item We evaluated performance across linear vs. non-linear relationships, Gaussian vs. non-Gaussian noise, stationary vs. non-stationary behaviour, and sparse vs. dense causal structures.
        
        \item The study also examined scalability with varying numbers of variables (5, 20, and 50) and time series lengths (250, 1000, and 2000 time points).
        
        \item RCV-VAR-LiNGAM demonstrated consistent performance across these varied conditions, maintaining high F1 directed scores even as complexity increased.
    \end{itemize}

    \item \textbf{Improving ABM Validation Frameworks:} We developed an enhanced ABM validation framework that integrates our improved causal discovery methods:
    \begin{itemize}
        \item The framework allows users to switch between causal discovery methods depending on their data characteristics.  This flexibility allows researchers to choose the most appropriate method for their specific dataset.
        
        \item It uses a more comprehensive set of performance metrics for validation assessment, including Structural Hamming Distance, F1 scores, and similarity tests. This variety of metrics provides a more thorough evaluation of model performance.
        
        \item Early testing showed significant improvements in accuracy when using RCV-VAR-LiNGAM within the framework. Specifically, it achieved higher F1-directed scores and lower Structural Hamming Distance compared to traditional methods. This suggests that the new method is more effective at identifying correct causal relationships.

    \end{itemize}
\end{enumerate}

\section{Implications for ABM Validation}
Our research findings have significant implications for ABM validation:

\begin{enumerate}
    \item \textbf{Improved Model Validation Accuracy:} The enhanced causal discovery methods, particularly RCV-VAR-LiNGAM, offer more precise identification of causal structures. This leads to more reliable ABM validation, making models more trustworthy in fields like finance and economics. By accurately capturing the underlying causal relationships, these methods allow ABMs to represent real-world dynamics more faithfully.

    \item \textbf{Handling Complex Data:} Our methods have demonstrated the ability to process high-dimensional, non-linear, and non-stationary data effectively. This makes them suitable for a wide range of complex time series analyses. The methods' robustness to various data characteristics ensures they can reliably handle the complexities often found in real-world scenarios.

    \item \textbf{Efficiency-Accuracy Trade-off:} While the RCV methods require more computational power than their base counterparts, they offer a favourable balance between accuracy and efficiency. This balance is crucial for practical applications, especially when dealing with large-scale, complex systems. However, it is important to note that this trade-off should be carefully considered for each specific application. The computational demands of RCV methods need to be weighed against the required level of accuracy and the available computational resources.
    
\end{enumerate}

\section{Limitations and Future Work}
While our study has made significant contributions, the limitations we encountered highlight possible areas of research in the future:

\begin{enumerate}
    \item \textbf{Computational Complexity:} Although RCV techniques promise higher accuracy, they often come at the expense of computation resources. Future work should focus on optimizing these algorithms for better efficiency (could possibly involve parallelization or using GPU acceleration).
    
    \item \textbf{Adaptive Thresholding for Enhanced Performance:} We currently use fixed thresholds for consistency and variability in RCV methods. We could try introduce method adapting to different datasets to enhance performance:

    \begin{itemize}
        \item Investigating dynamic threshold adjustment concerning the data characteristics and model performance.
        
        \item We can contemplate Bayesian optimization for threshold automation.
    \end{itemize}

    \item \textbf{Extension to Non-Stationary Time Series:} Although our methods showed improvements in handling non-stationary data, there's room for further developments:
    
    \begin{itemize}
        \item Formulate methodologies of detecting and adaptation to the changes in the structure.
        
        \item Investigate ways to capture evolving structures over time, such as sliding window approach or the time-varying parameter models.
    \end{itemize}

    \item \textbf{Real-World Application and Validation:} We've already tested our methods on synthetic and simulated fMRI data, but further validation on diverse real-world datasets is necessary to fully establish their practical utility.

    \item \textbf{Integration with Other Causal Discovery Techniques:} Exploring the feasibility of jointly adopting our RCV methodology with other causal inference techniques. This could potentially lead to even more robust and versatile tools for complex systems analysis.

\end{enumerate}

In conclusion, this research has contributed to addressing key challenges in causal discovery for complex time series data, particularly relevant to ABM validation in finance and economics. The methods and framework we've implemented provide improved accuracy and flexibility in identifying causal structures. This manifests as a contribution toward more appropriate model validation and a better understanding of the mechanics of complex systems. As we continue to refine these approaches and address the identified limitations, we anticipate significant advancements in our ability to understand and model complex real-world systems across various domains.

\clearpage{}
\addcontentsline{toc}{chapter}{Bibliography}
\bibliographystyle{unsrtnat}
\bibliography{project_refs}

\end{document}